\pdfoutput=1
\documentclass[10pt, a4paper, twocolumn]{naverlabseurope}
\usepackage{times}
\usepackage{latexsym}
\usepackage{booktabs}
\usepackage{graphicx}
\usepackage{enumitem}

\usepackage[T1]{fontenc}
\usepackage[utf8]{inputenc}

\usepackage{microtype}

\usepackage{inconsolata}

\newcounter{mycounter}

\newcommand\labelledmodelcounter[1]{\hspace{-0.5cm}\refstepcounter{mycounter}\textbf{(\themycounter)}\label{model:#1}}
\newcommand{\modelref}[1]{\textbf{(\ref{model:#1})}}

\title{Investigating the potential of Sparse Mixtures-of-Experts for multi-domain neural machine translation}

\authors{Nadezhda Chirkova \quad
Vassilina Nikoulina \quad
Jean-Luc Meunier \quad
Alexandre B\'erard
  }
\affiliations{NAVER LABS Europe}
\website{}
\websiteref{}
\contributions{}
\teaserfile{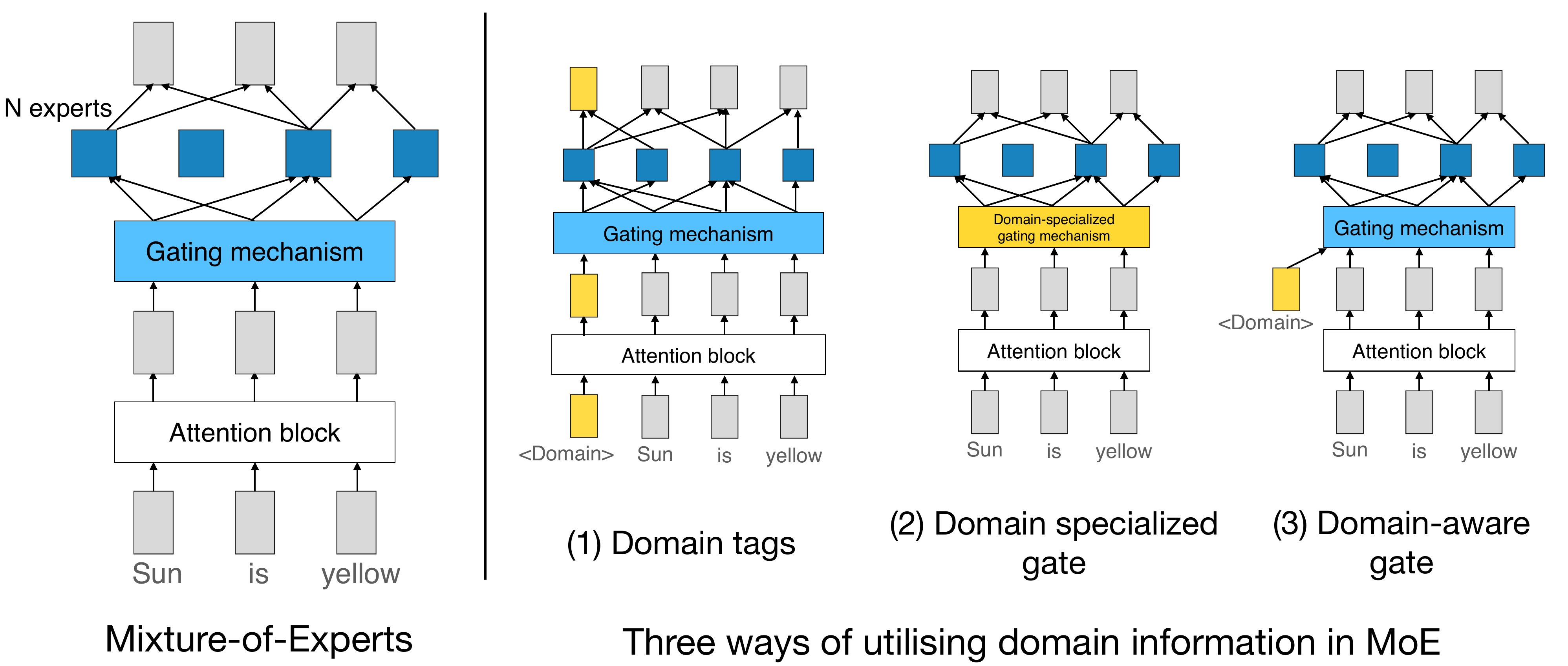}
\teasercaption{}
\correspondingauthor{[nadia.chirkova, vassilina.nikoulina]@naverlabs.com}

\begin{abstract}
We focus on multi-domain Neural Machine Translation, with the goal of developing efficient models which can handle data from various domains seen during training and are robust to domains unseen during training. We hypothesize that Sparse Mixture-of-Experts (SMoE) models are a good fit for this task, as they enable efficient model scaling, which helps to accommodate a variety of multi-domain data, and allow flexible sharing of parameters between domains, potentially enabling knowledge transfer between similar domains and limiting negative transfer. We conduct a series of experiments aimed at validating the utility of SMoE for the multi-domain scenario, and find that a straightforward width scaling of Transformer is a simpler and surprisingly more efficient approach in practice, and reaches the same performance level as SMoE. We also search for a better recipe for robustness of multi-domain systems, highlighting the importance of mixing-in a generic domain, i.e. Paracrawl, and introducing a simple technique, domain randomization.
\end{abstract}

\begin{document}
\maketitle
\section{Introduction} 

The goal of multi-domain Neural Machine Translation (NMT) is to build a system capable of generating high-quality translations for inputs from various domains. Such domains are given in advance and are represented by different sources of data, eg. European Parliament speeches, scientific articles, etc. %Domains could also include a \textit{generic} domain in order to make the model more robust to out-of-domain examples. 
At the test time, input examples may come with or without known \textit{domain labels}, depending on the practical scenario.

Most popular approaches to equip multi-domain NMT model with domain knowledge do so either via \textit{domain tags} \citep{kobus-etal-2017-domain,britz-etal-2017-effective,berard-etal-2019-machine} or \textit{domain adapters} \citep{bapna-firat-2019-simple}. When domain knowledge is integrated via domain tags, prepended to input sequences as special tokens, the model is able to benefit fully from knowledge transfer between domains via shared representations. However, the model may suffer from negative transfer if it does not have enough parameters to accommodate contradictory signals from different domains. Domain adapters provide a separate set of parameters per each domain, therefore increasing the model size and being robust to negative transfer, but may not benefit from potential knowledge transfer across different domains.

In this work, we investigate the potential of Sparse Mixture-of-Experts models (SMoE, \citealp{shazeer2017}) for multi-domain machine translation. SMoE belongs to a family of conditional-computation models,  and uses a gating mechanism to decide which subset of model parameters will be activated for each input token. SMoE seems to be a good fit for multi-domain scenario due to several reasons. First, SmoE scales the model size, which appears to be necessary for a variety of multi-domain data the model needs to accommodate, but keeps constant the inference floating-point operations (FLOPs), thus enabling \textit{efficient scaling}. Second, SMoE looks as a reasonable and 
\textit{effective middle ground} between domain tags and domain adapters, due to the enabled ``soft'' sharing of parameters between domains. In particular, SMoE has a potential to learn which domains should be processed with the same set of parameters, i.e. to enable knowledge transfer between related domains, and which domains should not share parameters, i.e. to prevent negative transfer. To fully benefit from the domain information when it is available, we study three ways of utilizing domain labels in SMoE.
We conduct series of experiments, aiming to validate \textit{the hypothesis that SMoE is an effective and efficient architecture for the multi-domain scenario}. Our experiments lead us to \textit{non-conclusive results} regarding the utility of the SMoE architecture, but highlight the \textit{advantages of model scaling}, particularly straightforward width scaling, in both effectiveness and inference time efficiency.

Our findings can be summarized as follows:
\begin{itemize}[noitemsep,topsep=0pt,parsep=0pt,partopsep=0pt]

\item \textit{Effectiveness.} SMoE significantly outperforms the baseline Transformer Base architecture, mostly due to the model scaling effect: SMoE performs on par with the horizontally scaled model of the same total size (width scaling of the Transformer dimension);
\item \textit{Efficiency.} For models that fit into a single GPU, a straightforward \textit{width scaling} of the Transformer dimension introduces almost no computational overhead compared to the non-scaled model, when modern GPUs such as Tesla A100 or V100 are used. On the contrary, SMoEs do introduce a substantial computational overhead at inference. As a result, the horizontally scaled model turns out to be an efficient scaling approach.
\item \textit{Domain knowledge in SMoE.} We introduce and compare several methods for incorporating domain knowledge into SMoE and find that that the simplest approach of using \textit{domain tags} performs similarly to the more advanced approaches (and to the horizontally scaled model also equipped with domain tags);
\item \textit{Robustness.} We search for a recipe for more robust multi-domain systems, i.e. robust to out-of-domain examples and wrong domain labels, and highlight two important components. First, we show that mixing-in a generic domain improves model performance on out-of-domain examples. Second, we introduce a simple technique, \textit{domain randomization}, targeted at enriching \textit{out-of-domain} robustness. We observe that it also substantially increases robustness to wrong domain labels.

\end{itemize}

\section{Related work}

\paragraph{Multi-domain Neural Machine Translation.} In Multi-domain NMT we assume that our training data comes from different sources, either with well identified domains, or more generic ones (eg. web-crawled data). The goal is to exploit this knowledge efficiently during training and/or inference.  \citet{kobus-etal-2017-domain,britz-etal-2017-effective,tars2018multidomain,berard-etal-2019-machine} specify domain information via special \textit{domain tags} prepended to the source or target sequence.  Thus the model has access to domain information by accessing this domain tag via the attention mechanism, and its parameters are shared across all domains. Several works \citep{jiang-etal-2020-multi-domain,dabre2020comprehensive,britz-etal-2017-effective} have tried to decouple domain-specific representations from domain independent representations by proposing different architectural changes and/or adversarial training.

Most of the above mentioned approaches to multi-domain NMT are only interested in optimizing the performance on the domains observed during training and do not consider the performance of such models on new previously unseen domains.  Moreover, these works assume that the domain is well identified at inference time, while in real life serrings NMT model may have to deal with wrong or unspecified domain label, or mixed-domain inputs.

%\cite{aharoni-goldberg-2020-unsupervised}
\citet{pham-etal-2022-multi,wang-etal-2020-learning-multi} propose domain resampling strategies attempting to maximize performance across all the domains, and acting as a regularization to improve robustness to unseen domains. We address robustness from another (complementary) perspective, by including a generic domain in the training data and by explicitly encouraging models' reliance on generic data. We also study a wider and more diverse set of unseen domains.
 \citet{pham-etal-2021-revisiting} has performed and extensive benchmark of existing multi-domain NMT approaches by explicitly evaluating their \textit{Effectiveness} (benefit from domain knowledge availability  and  benefit from domain transfer between training domains)
 and \textit{Robustness} to the fuzzy or unknown domains at inference time. Authors concluded that none of the existing approaches is competitive with simply training a model on a mixture of all datasets and further finetuning on each domain's dataset, which implies that in existing approaches the impact of negative transfer across domains is higher than benefit of the positive transfer between training domains.

\paragraph{Mixture of Experts models for Neural Machine Translation.}
The Mixture-of-Experts architecture has been shown to be an efficient way for models parameters scaling \cite{glam,kim2021scalable}.  It has demonstrated promising results in NMT for multilingual settings \citep{flores,kim2021scalable,kudugunta-etal-2021-beyond-distillation}, where efficient scaling of the parameters allowed to cope with the \textit{curse of multilinguality} while keeping reasonable inference cost.
To the best of our knowledge, the potential of SMoE architecture has not been explored in multi-domain NMT. 
\citet{pham-etal-2022-latent} propose an approach which learns the selection of domain-specific parameters end-to-end so that domains can share a part of parameters. The SMoE architecture in fact generalizes this idea further by enabling the selection of parameters specific to a combination of a token and a domain.

In this work we study whether Sparse Mixture of Experts architectures are beneficial in multi-domain NMT settings, and to what extent domain information is important when we scale up the model.

\section{Methodology}
\label{sec:methods}

Following~\cite{pham-etal-2021-revisiting}, we define domains as different sources of data used to train a machine translation model. The \textit{effectiveness} of the model is defined as the performance of the model on the \textit{seen} domains, i.e. the domains included in the training data.
We also evaluate \textit{out-of-domain robustness} of the model, i.e. the performance on the domains \textit{unseen} during training, i.e. on some held-out test domain data from different data distribution than the training domains.  At the same time, we care about model \textit{efficiency}, i.e. inference speed, as it is crucial in production systems.
 
To improve out-of-domain robustness, we include a large generic corpus in the training data (Paracrawl, \cite{paracrawl}). 

We consider two practical scenarios. The first scenario is the most common in the literature, in which each test sentence comes with the known domain tag. This scenario often arises in b2b settings,  e.g. a cloud provides a translation service to other companies knowing which test data is from which company.  In the second scenario, more common in online b2c services, the users do not provide explicit domain information, and the model should be able to handle various inputs without knowing their domains.

We attempt to gain a better understanding on the following questions: 
\begin{itemize}[noitemsep,topsep=0pt,parsep=0pt,partopsep=0pt]
    \item how does SMoE perform in terms of \textit{ effectiveness}, \textit{out-of-domain robustness} and \textit{efficiency}; 
    \item what is the best strategy for domain knowledge integration in SMoE;
    \item whether the knowledge about the (seen) domains benefits scaled model performance on these domains;
    \item how we can improve out-of-domain robustness further.
\end{itemize}

\subsection{Mixture of Experts} 
\label{sec:methods_SMoE}
Sparsely-gated Mixture-of-Experts (SMoE) models activate a \textit{subset} of their parameters \textit{per input token}, contrary to dense models, where the \textit{entire network is used for each input token}. Therefore, the total amount of parameters can be significantly increased with limited impact on the computational cost. 

In the SMoE Transformer models proposed by \citet{lepikhin2020gshard}, the FFN sublayers in the dense model are replaced with SMoE layers consisting on $N$ experts each (with the same architecture as an FFN layer). A SMoE layer takes an input token representation $x_t$ and then routes it to the top~$k$ out of $N$ experts according to a learned gating mechanism, which computes a probability distribution over all the experts: 
$G_t = softmax(W_g \cdot x_t)$.

The output of the SMoE layer is a weighted sum of the outputs of the top $k$ selected experts $\mathcal{E}$:
\begin{equation}
    y_t = \frac{1}{\sum_{i \in \mathcal{E}} G_{t,i}} \sum_{i \in \mathcal{E}} G_{t,i}E_{i}(x_t)
\end{equation}
In addition to the efficient scaling, the SMoE architecture may provide more flexibility in multi-domain settings: if it is able to specialize experts for different domains it would lead to lower negative transfer between training domains. 

For fair evaluation, we compare the performance of the SMoE model with two fully dense models that either match its total number of parameters or its FLOPs at inference (more details in section \ref{sec:exp_details}).

\begin{figure}[h!]
    \centering
         \includegraphics[width=\linewidth]{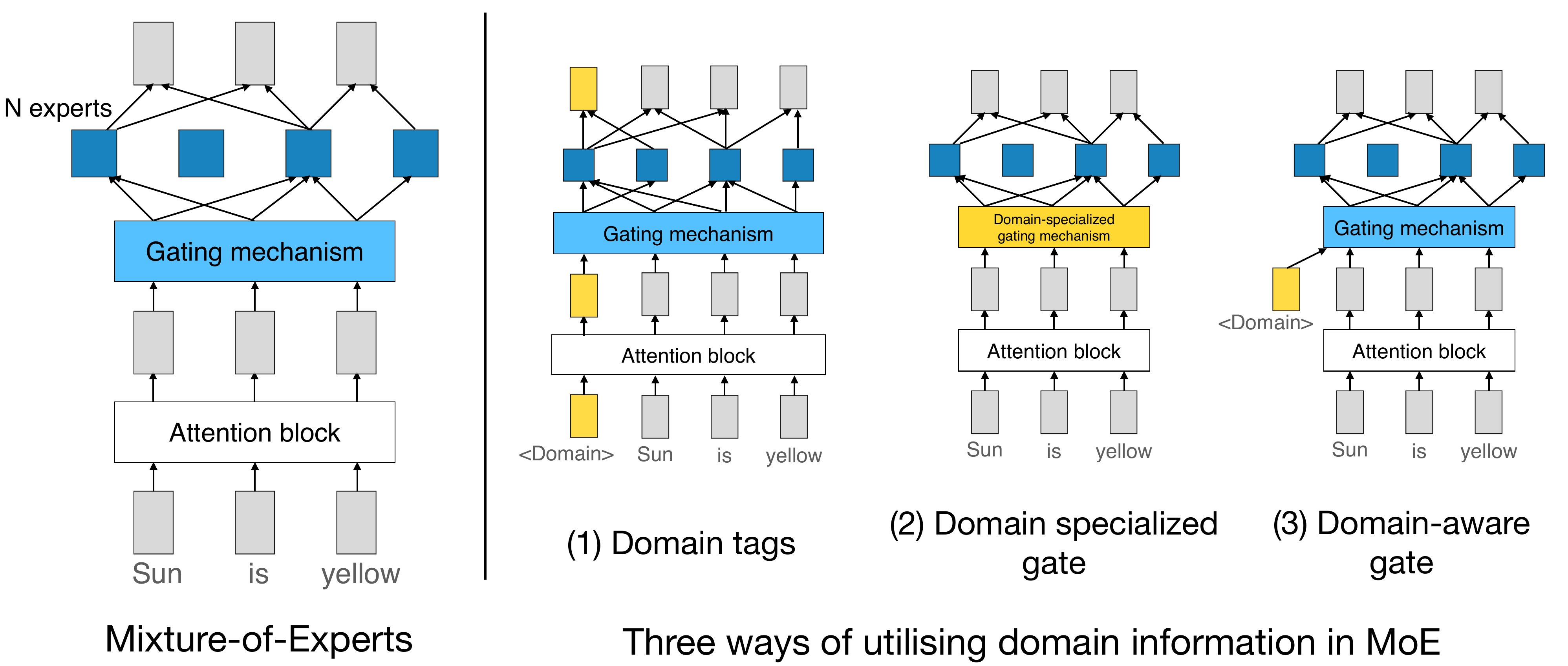} 
        \caption{Illustration of Mixture-of-Experts (left) and three considered ways of utilizing domain information in Mixture-of-Experts (right).}
        \label{fig:illustration}
\end{figure}

\subsection{Domain Knowledge integration} Domain knowledge is presented to the model in the form of a 
%learnable domain embedding $e_{d}$. 
domain label $d$.
We compare different ways of integrating such
domain label
%embedding $e_d$
into the SMoE model, see their illustration in Figure~\ref{fig:illustration}.
\paragraph{Domain Tags} prepend domain embedding $e_d$ to the source sequence similar to \cite{kobus-etal-2017-domain,britz-etal-2017-effective}.  In this settings domain embeddings represent the only domain-specific parameters, the rest of the parameters are shared across all the domains. The model attends to domain information via self- and cross-attention during training and inference and modifies its internal representations accordingly.
\paragraph{Domain Specialized Gate} \citet{gupta2022sparsely} introduce separate task-specialized gates in SMoE for multitask training of language models with the goal to encourage different experts to specialize to different tasks, while limiting the negative transfer between tasks. Similarly, we consider domain-specific gates $G^d_t$ that devote specialized gating mechanism parameters ($W^d_g$) to each domain. In multi-domain settings, such specialization is expected to encourage the model to a select different set of experts for different domains.
\paragraph{Domain-aware Gate.} A fully domain-specialized gating mechanism may limit knowledge transfer between different domains. Moreover, learning gating mechanism parameters can be more challenging for very small domains, with very small amount of samples in the training data. Therefore we propose a \textit{Domain-aware gating mechanism} as an alternative to the domain-specialized gate. It takes as input the concatenation of the current token representation $x_t$ and the domain embedding $e_d$:
\begin{equation}
    G_t(d) = softmax(W_g \cdot (x_t \oplus e_d))
\end{equation}

\subsection{Domain randomization}

As discussed above, our training data includes several given domains, as well as a large generic domain.

In practice, we would like to rely on the \textit{generic} domain to translate text from unseen domains. In order to make the \textit{generic} domain representation more robust to unseen domains, we introduce \textit{domain randomization} during training. 

It consists in randomly assigning training examples from well-defined domains to the \textit{generic} domain, with some probability $p$. In our experiments, we use $p=0.5$.
We expect such randomization will allow the model to associate the generic tag with richer data samples and particularly to recognize situations when a test example is close to domains seen during training.

By introducing domain randomization, we assume that using only a half of sampled data points from a domain is enough to train a small amount of domain-specific parameters (domain tags, embeddings, or domain-specific gating mechanism), and our experiments confirm this assumption: using domain randomization improves robustness without hurting performance on seen domains. Another assumption is that the model capacity is sufficient to accommodate all the seen domains and at the same time associate the generic domain with all the seen domains. This assumption is well aligned with the suggested model scaling.

\section{Experimental setup}
\label{sec:exp_details}
\paragraph{Data.}
We experiment with the German-English (De-En), English-German (En-De) and English-French (En-Fr) translation directions. 
All models are trained on a mix of generic data (Paracrawl, \citealp{paracrawl}) and several \textit{seen} domains: "Law", "Medical", "Ted", "Subtitles", and "Patents".
The selection of seen domains is motivated by having diverse domains of different sizes and having a pair of close domains, "Ted"  and "Subtitles" (they both relate to spoken language). We also test the generalization capabilities of models on a set of \textit{unseen} domains: "Flores", "IT", "Koran", "Patents (medical)", "Subtitles (Medical)", and "Medical abstracts".
The purpose of introducing the last three unseen domains is to test models' capabilities to recognize unseen domains related to some seen domains ("Medical", "Patents", and "Subtitles"). 

We deduplicate train-test splits to avoid data contamination, i.e. that test data does not leak into training data. We provide further details about  datasets in the Appendix~\ref{app:exp_setup}.

\paragraph{Model hyperparameters.}
% introduce this setting above, mention knowledge transfer somewhere
For De-En and En-Fr translation, we use standard Transformer Base hyperparameters from~\citet{vaswani}: 6 encoder layers and 6 decoder layers, 8 heads, $d_{model}=512$, and $d_{ff}=2048$. For En-De, we test the applicability of SMoE for a shallow configuration with only 3 decoder layers, which was shown to be much faster at inference, with a negligible performance drop~\cite{shallow, berard-etal-2021-efficient}. SMoE model uses 10 experts every second layer with top-2 token-level gating. Training details are given in Appendix~\ref{app:exp_setup}.

\paragraph{Models.}
We consider two settings of multidomain NMT models: with access to the labels of seen domain and without it.
 In the setting where domain labels are not used, we compare the commonly-used Transformer Base~\cite{vaswani}, SMoE, and two additional dense variants: Transformer width $\times$1.5 (matches SMoE inference FLOPs SMoE)\footnote{Since we place experts every second layer, the width only needs to be increased 1.5 times to reach the FLOPs of SMoE.} and Transformer width $\times$5 (matches total parameter count of SMoE).
By increasing width we mean increasing $d_{ff}$.

In the setting with domain labels being available during both training and inference, we compare three ways of utilizing domain information in SMoE (see Section~\ref{sec:methods}) with and without proposed domain randomization, and two commonly used baselines: Transformer with domain tags and Transformer with domain adapters. We treat domain adapters as domain-specific parameters and train them together with the rest of model parameters. We set the adapter dimension to $2048$, to match the number of inference FLOPs and parameters with SMoE. The number of parameters is matched because the number of experts we use equals the number of seen domains. % and Transformer trained with domain-specific domain adapters. 

\paragraph{Evaluation.} For each model, we report BLEU scores measured with SacreBLEU in the main text and COMET scores in Appendix, as well as inference FLOPs and the number of parameters. To estimate the number of floating point operations (FLOPs), we count matrix multiplications in the forward pass, assuming source and target sequence lengths of 10.

\begin{table*}[ht!]
\normalsize
\centering
\begin{small}
%\begin{tabular}{cp{1cm}p{1cm}p{1cm}p{1cm}p{1cm}p{1cm}}
\begin{tabular}{p{5.cm}cc|ccccc|c}
\toprule
\textbf{Model} & \textbf{\#Params} & \textbf{FLOPs} & \textbf{Law} & \textbf{Med} &   \textbf{Ted} & \textbf{Subt} & \textbf{Pat} &  \textbf{Avg} \\
\midrule
\multicolumn{9}{c}{Without access to seen domains labels} \\ \midrule

\labelledmodelcounter{TB} Transformer Base (TB)   & 56M &   0.44B        &  52.0 &    47.5 &  \underline{40.9} &      {29.1} &    {31.9} &  40.3 \\

\labelledmodelcounter{Tw2} Transformer width $\times$1.5$^*$   & 69M   &   0.57B     &  52.3 &    48.0 &  41.3 &      {29.5} &    {31.8} &  40.6 \\
\labelledmodelcounter{Tw5} Transformer width $\times$5     & 160M   & 1.44B     &  \textbf{54.7} &    \textbf{49.4} &  \textbf{42.1} &      \textbf{30.3} &    \textbf{32.9} &  \textbf{41.9} \\
\labelledmodelcounter{SMoE} Mixture of Experts (SMoE)  &  170M   &  0.57B   &  \textbf{54.5} &    \textbf{49.8} &  \textbf{41.8} &      {29.9} &    \textbf{32.4} &  \textbf{41.7} \\
\midrule
\multicolumn{9}{c}{With access to seen domains labels} \\ \midrule
\labelledmodelcounter{TBtags} TB + domain tags         & 56M &   0.44B          &  \underline{52.1} &    47.7 &  \underline{42.3} &      {29.5} &    \underline{32.0} &  \underline{40.7} \\
\labelledmodelcounter{TBtagsDR} TB + domain tags + DR          & 56M &   0.44B      &  \underline{52.1} &    47.7 &  \underline{42.3} &      {29.6} &    \underline{31.9} &  \underline{40.7} \\
\labelledmodelcounter{TBtagsw2} Tr. $\times$1.5 + domain tags$^*$     & 69M   &   0.57B    &  {52.7} &    {48.1} &  {42.7} &      {29.7} &    {32.4} &  41.1 \\
\labelledmodelcounter{TBtagsw2DR} Tr. $\times$1.5 + domain tags + DR        & 69M &   0.57B        &  52.9 &    48.2 &  42.5 &      29.5 &    \underline{32.0} &  41.0 \\
\labelledmodelcounter{Tw5tags} Tr. $\times$5 + domain tags        & 160M  &   1.44B          &  \textbf{54.8} &    \textbf{50.0} &  43.0 &      \textbf{30.7} &    32.3 &  \textbf{42.2}  \\
\labelledmodelcounter{Tw5tagsDR} Tr. $\times$5 + domain tags + DR          & 160M  &   1.44B      &  \textbf{54.5} &    \textbf{49.9} &  \textbf{43.2} &      \textbf{31.3} &    32.7 &  \textbf{42.3} \\
\labelledmodelcounter{TBadapters} TB + domain adapters       & 170M   &   0.57B      &  \textbf{55.1} &    \textbf{50.8} &  \underline{38.3} &      \underline{28.2} &    32.3 &  41.0 \\
\labelledmodelcounter{TBadaptersDR} TB + domain adapters + DR       &  170M  &   0.57B  &  52.6 &    48.0 &  \underline{40.6} &      \underline{28.8} &    \underline{32.0} &  \underline{40.4} \\
\labelledmodelcounter{SMoEtags} SMoE + domain tags        &  170M   &  0.57B      &  \textbf{54.4} &    \textbf{50.0} &  \textbf{43.3} &      \textbf{30.4} &    32.2 &  \textbf{42.1} \\
\labelledmodelcounter{SMoEtagsDR} SMoE + domain tags + DR      &  170M   &  0.57B    &  \textbf{54.2} &    \textbf{50.4} &  \textbf{43.3} &      \textbf{30.7} &    32.2 &  \textbf{42.1} \\
\labelledmodelcounter{DaSMoEv1} SMoE + Dom. Aware Gate      &  170M   &  0.57B     &  \textbf{54.9} &    \textbf{50.5} &  \textbf{43.1} &      30.3 &    32.4 &  \textbf{42.2} \\
\labelledmodelcounter{DaSMoEv1DR} SMoE + Dom. Aware Gate + DR     &  170M   &  0.57B  &  \textbf{54.5} &    \textbf{50.2} &  \textbf{43.2} &      \textbf{30.9} &    32.2 &  \textbf{42.2} \\
\labelledmodelcounter{DaSMoEv2} SMoE + Dom. Spec. Gate     &  170M   &  0.57B &  \textbf{54.6} &    \textbf{50.3} &  42.7 &      30.1 &    32.5 &  \textbf{42.0} \\ 
\labelledmodelcounter{DaSMoEv2DR} SMoE + Dom. Spec. Gate + DR     &  170M   &  0.57B &  \textbf{54.8} &    \textbf{49.8} &  \textbf{43.1} &      \textbf{30.6} &    32.4 &  \textbf{42.2} \\
\bottomrule
\end{tabular}
\end{small}
\caption{BLEU scores for seen domains, German-English translation. DR: domain randomization. \textbf{Bold} denotes significant improvement vs. "Transformer width $\times$1.5" (without domain labels) or "Tr. $\times$1.5 + domain tags" (with domain labels), both marked with $^*$; \underline{underline} denotes significant losses.}
\label{tab:main_seen}
\end{table*}

\begin{table*}[ht!]
\normalsize
\centering
\begin{small}

\begin{tabular}{p{5.5cm}|cccp{1cm}p{1cm}p{1cm}|c}
\toprule
\textbf{Model}   & \textbf{Flores} & \textbf{IT} & \textbf{Koran} & \textbf{Pat (med)} & \textbf{Subt (med)} & \textbf{Med abst} &  \textbf{Avg} \\
\midrule
\multicolumn{8}{c}{Without access to seen domains labels} \\ \midrule
\textbf{(1)} Transformer Base (TB)      &      {41.0} &  34.4 &  14.0 &              {35.9} &                {37.4} &              16.7 &  29.9 \\

\textbf{(2)} Transformer width $\times$1.5$^*$     &      {41.0} &  34.4 &  {13.9} &              {36.0} &                {37.3} &              16.6 &  29.9 \\
\textbf{(3)} Transformer width $\times$5       &      \textbf{41.7} &  34.7 &  \textbf{14.5} &              \textbf{36.4} &                \textbf{38.7} &              16.7 &  \textbf{30.4} \\
\textbf{(4)} Mixture of Experts (SMoE)   &      \textbf{41.9} &  34.7 &  \textbf{14.7} &              {35.9} &                \textbf{38.3} &              16.8 &  \textbf{30.4} \\
\midrule
\multicolumn{8}{c}{With access to seen domains labels} \\ \midrule
\textbf{(5)} TB + domain tags           &      \underline{40.4} &  {33.8} &  {13.7} &              {34.7} &                \underline{34.7} &              16.9 &  29.0 \\
\textbf{(6)} TB + domain tags + DR      &      \underline{40.4} &  \underline{32.5} &  {13.6} &              \textbf{35.8} &                \textbf{36.8} &              \underline{16.6} &  29.3 \\
\textbf{(7)} Tr. $\times$1.5 + domain tags$^*$     &      {41.3} &  {33.8} &  {13.9} &              {34.6} &                {35.6} &              {17.2} &  29.4 \\
\textbf{(8)} Tr. $\times$1.5 + domain tags + DR     &      41.0 &  33.9 &  14.0 &              \textbf{36.0} &                \textbf{38.0} &              \underline{16.8} &  \textbf{29.9} \\
\textbf{(9)} Tr. $\times$5 + domain tags          &      41.6 &  34.2 &  \textbf{14.9} &              \textbf{35.2} &                35.6 &              18.0 &  \textbf{29.9} \\
\textbf{(10)} Tr. $\times$5 + domain tags + DR    &      41.6 &  \textbf{34.6} &  \textbf{14.6} &              \textbf{36.2} &                \textbf{38.2} &              \underline{16.9} &  \textbf{30.4} \\
\textbf{(11)} TB + domain adapters       &      \underline{40.5} &  {34.0} &  \textbf{14.2} &              \underline{31.9} &                \underline{33.3} &              17.0 &  \underline{28.5} \\
\textbf{(12)} TB + domain adapters + DR  &      {41.1} &  {33.7} &  \textbf{14.3} &              \textbf{36.1} &                \textbf{37.6} &              \underline{16.8} &  \textbf{29.9} \\
\textbf{(13)} SMoE + domain tags          &      {41.9} &  \textbf{34.9} &  \textbf{14.9} &              {34.8} &                {35.8} &              17.3 &  \textbf{29.9} \\
\textbf{(14)} SMoE + domain tags + DR     &      {41.8} &  \textbf{34.7} &  \textbf{15.4} &              \textbf{36.1} &                \textbf{38.2} &              17.1 &  \textbf{30.6} \\
\textbf{(15)} SMoE + Dom. Aware Gate      &      {41.6} &  \textbf{35.2} &  \textbf{14.6} &              \underline{31.9} &                {35.9} &              \underline{16.7} &  29.3 \\
\textbf{(16)} SMoE + Dom. Aware Gate + DR &      {41.8} &  \textbf{35.1} &  \textbf{14.8} &              \textbf{36.0} &                \textbf{38.1} &              \underline{16.5} &  \textbf{30.4} \\
\textbf{(17)} SMoE + Dom. Spec. Gate &      41.6 &  \textbf{35.0} &  \textbf{14.6} &              \underline{30.0} &                35.7 &              16.8 &  28.9 \\ 
\textbf{(18)} SMoE + Dom. Spec. Gate + DR &      {41.3} &  \textbf{34.7} &  \textbf{14.8} &              \textbf{36.2} &                \textbf{38.0} &              \underline{16.8} &  \textbf{30.3} \\
\bottomrule
\end{tabular}
\end{small}
\caption{BLEU scores for unseen domains, German-English translation. For models with access to seen domain labels (second group of rows), the ``generic'' label is used for unseen domains. DR: domain randomization. \textbf{Bold} denotes significant improvement vs. "Transformer width $\times$1.5" (without domain labels) or "Tr. $\times$1.5 + domain tags" (with domain labels), both marked with $^*$; \underline{underline} denotes significant losses.}
\label{tab:main_unseen}
\end{table*}

\section{Results}

\subsection{Study of scaling methods}

Tables~\ref{tab:main_seen} and~\ref{tab:main_unseen} report German-English results for seen and unseen domains respectively. In the Appendix, we present detailed results in the other translation directions (En-De and En-Fr) with the same conclusions.   
Table~\ref{tab:all_directions} summarizes results for these translation directions. 
\paragraph{Comparison of scaling approaches.} 

We observe that \textit{the performance of SMoE is close to the performance of dense model with the same parameter count} (Transformer width x5), on both seen and unseen domains, in both settings with and without access to seen domains labels. Here we compare models \modelref{Tw5} vs \modelref{SMoE} and \modelref{Tw5tags} vs \modelref{SMoEtags} which correspond to different ways to increase model capacity. Compared to Transformer Base, they both bring 
%0.2--
up to 2.5 BLEU improvement (1.4 BLEU on average) on seen domains and 
%0-
up to 1.3 BLEU improvement (0.6 BLEU on average) on unseen domains. Transformer width $\times$1.5 (which serves as a baseline for SMoE, with the same inference FLOPs) performs only slightly better than Transformer Base, i.e. \modelref{Tw2} vs \modelref{TB} and \modelref{TBtagsw2} vs \modelref{TBtags}.

Domain adapters \modelref{TBadapters} are capable of reaching the performance of SMoE with tags \modelref{SMoEtags} or width scaling with tags \modelref{Tw5tags} on some seen domains ("Law" and "Medical") but perform significantly worse than Transformer Base with tags \modelref{TBtags} on other seen domains ("Ted" and "Subtitles"). This happens because of the uneven convergence speed of different adapters: for the "Ted" and "Subtitles" domains, validation performance starts to decrease at some point during training (we did not observe this behaviour for any other models in our experiments). For unseen domains, domain adapters rely on \textit{generic} adapter 
 and usually perform on a par with Transformer Base with tags \modelref{TBtags}, i.e. worse than scaled models. The underlying problem behind these two observations is that adapters allocate the same capacity to all domains, including the generic domain adapter used to process unseen domains.\footnote{Though it is in principle possible to use individual adapter dimensions for different domains, it is unclear how to choose these hyperparameters in practice.} In contrast, scaled models can intrinsically allocate variable capacity to each domain, at the same time enabling implicit knowledge transfer between domains.

\paragraph{Utilization of domain information in SMoE.} 
We will now focus only on a setting with available domain information (bottom parts of tables~\ref{tab:main_seen} and~\ref{tab:main_unseen}). 
Comparing three ways of utilizing domain information in SMoE, described in Section~\ref{sec:methods_SMoE}, we observe that \textit{the simplest approach with domain tags \modelref{SMoEtags} performs similarly to more complex approaches}, which use domain embeddings in the gating mechanism, SMoE + Dom.Aware Gate, \modelref{DaSMoEv1} or domain-specific gating parameters~\citep{gupta2022sparsely}, SMoE + Dom.Spec. Gate, \modelref{DaSMoEv2}. This holds for both seen and unseen domains.

\paragraph{Are domain tags helpful?} When comparing models with and without domain tags, e.g. \modelref{TBtags} vs \modelref{TB}, \modelref{TBtagsw2} vs \modelref{Tw2}, \modelref{Tw5tags} vs \modelref{Tw5}, or \modelref{SMoEtags} vs \modelref{SMoE}, we observe \textit{an average BLEU gain of 0.5 on seen domains when domain labels are given}. On unseen domains, models equipped with domain tags perform mostly similarly to model without tags, except Transformer Base, where the domain tags can be detrimental. 
\begin{table}[ht!]
\normalsize
\centering
\begin{small}
%\begin{tabular}{cp{1cm}p{1cm}p{1cm}p{1cm}p{1cm}p{1cm}}
\begin{tabular}{p{2.4cm}|p{0.5cm}p{0.5cm}|p{0.5cm}p{0.5cm}}
\toprule
&  \multicolumn{2}{c|}{\textbf{En-De (6e+3d)}} &  \multicolumn{2}{c}{\textbf{En-Fr (6e+6d)}} \\
\textbf{Model / Domains} & \textbf{Seen} & \textbf{Unseen} & \textbf{Seen} & \textbf{Unseen} \\
\midrule
 
TB          &  35.3    &  25.2     &   45.5    &  31.3 
   \\   
%\textbf{(3)} 
Tr. $\times$5   &  36.6      &   26.0     &    47.0    &  32.2   \\
%\textbf{(4)} 
SMoE &   36.5    &   25.6    &   46.9    &     32.0      \\
\midrule
%\textbf{(6)} 
TB + tags         &  35.6     &  24.6    &   45.5    &    30.6    \\

Tr.$\times$5 +tags+DR        &  36.9  &  26.0      &  46.9   &  32.3 \\
SMoE+tags+DR   &  36.7   &   25.6    &  47.1   &      32.2      \\

\bottomrule
\end{tabular}
\end{small}
\caption{Summary of BLEU results, averaged over seen and unseen domains, for three translation directions. Detailed per-domain tables are given in Appendix.}
\label{tab:all_directions}
\end{table}

\paragraph{Effect of domain randomization.} 
We observe that \textit{domain randomization is an important component in out-of-domain robustness and is particularly helpful of unseen domains that are related to seen domains}. Models with access to seen domains labels but without domain randomization \modelref{Tw5tags} \modelref{SMoEtags} \modelref{DaSMoEv1} \modelref{DaSMoEv2} exhibit 1.3--2.2 BLEU drop on ~"Patents (medical)"~ and ~"Subtitles (medical)"~, because they process these domains with the ``generic'' tag which was never used for "Patents" or "Subtitles" data during training. With domain randomization \modelref{Tw5tagsDR} \modelref{SMoEtagsDR} \modelref{DaSMoEv1DR} \modelref{DaSMoEv2DR}, the ``generic'' tag is trained to work well with all seen domains and is thus robust to related unseen domains, i.e. performance on ~"Patents (medical)"~ and ~"Subtitles (medical)"~ is the same as of \modelref{Tw5} and \modelref{SMoE}. 
Importantly, the introduction of domain randomization does not hurt performance on seen domains.

For domain adapters, the use of DR \modelref{TBadaptersDR} leads to low performance on seen domains, because they have a lot of domain-specific parameters which are trained poorly with only a half of the domain data samples. For Transformer Base with tags, the use of DR \modelref{TBtagsDR} brings lower improvements on unseen domains related to seen domains, because of limited model capacity.

\begin{table}[]
    \centering
    \begin{small}
    \begin{tabular}{p{2.3cm}|p{1cm}p{1cm}p{1cm}}
    \toprule
     \textbf{Batch size (GPU)} & \textbf{TB}   &  \textbf{Tr. x5}  &  \textbf{MoE}  \\ \midrule
     1~~~~~~~~~~~~~~~(P40) &  3562.9  &  \textbf{3666.3}  &    6697.4	     \\ 
     1k  & 	128.5	  &  \textbf{144.1}  &    176.8	     \\ 
     10k  &  87.8	 &	\textbf{102.0}  &   	 \textbf{101.0}     \\ 
     100k  &  104.1    &  125.6  &   	\textbf{119.1}     \\ 
     \midrule
     %Batch size (GPU) & TB   &  Tr. x5  &  MoE  \\ \midrule
     1~~~~~~~~~~~~~~~(V100) &  1257.7  &  \textbf{1268.0}  &    1513.8     \\ 
     1k  & 94.4   &  \textbf{99.4}  &    121.5	     \\ 
     10k  &  27.8 &	\textbf{30.9}  &   34.2	      \\ 
     100k  &  37.2	  &  \textbf{38.5}  &    44.0	     \\ 
     \midrule
     %Batch size (GPU) & TB   &  Tr. x5  &  MoE  \\ \midrule
     1~~~~~~~~~~~~~~~(A100) &  864.8  &  \textbf{859.0}  &    1085.6     \\ 
     1k  &  80.2 &  \textbf{81.1}  &    92.4		     \\ 
     10k  &  20.8 &	\textbf{22.1}  &   	25.8      \\ 
     100k  &  29.3	  &  \textbf{32.3}  &   35.2		     \\ 
     \bottomrule
    \end{tabular}
    \end{small}
    \caption{Inference time (sec) for translating the Patents test set. Bold denotes fastest between "Tr. x5" and MoE.}
    \label{tab:inference_time}
\end{table}

\paragraph{Inference speed.} In Table~\ref{tab:inference_time} we report time measurements for translating the En-Fr "Patents" test set, averaged over three runs, for various Tesla GPUs. For SMoE, we choose the minimal time between our implementation and \verb|tutel| implementation~\cite{tutel}. We surprisingly find that for modern GPUs such as Tesla V100 or A100, which are highly optimized for matrix multiplications, increasing the model width leads to only marginal inference slow down. At the same time, SMoE is slower than "Tr. x5", despite 2.5x theoretical  FLOPs reduction compared to the wide model. In practice it is hard to implement SMoE in a way that it utilizes available GPU parallelism as efficiently as simple width scaling. However, for older GPUs, such as Tesla P40, with larger batch sizes, the gap between TB and "Tr. x5" inference times is higher (20\%) and SMoE implementation with \verb|tutel| can be compared to "Tr. x5". 
\begin{table}[]
\centering
\begin{small}
\begin{tabular}{l|p{0.45cm}p{0.45cm}p{0.45cm}p{0.45cm}p{0.45cm}}
\toprule
{} &  \textbf{Subt} &    \textbf{Ted} &    \textbf{Law} &  \textbf{Med} &  \textbf{Pat} \\
\midrule
MoE + tags       &      30.5 &  29.3 &  23.2 &    25.5 &  24.7 \\
MoE + tags + DR  &      30.7 &  30.0 &  28.7 &    30.3 &  30.1 \\
MoE Dom.aware gate      &      30.3 &  29.1 &  23.0 &    21.5 &  23.0 \\
MoE Dom.aw. + DR &      31.0 &  30.1 &  28.0 &    27.1 &  29.7 \\
MoE Dom.spec gate      &      30.1 &  29.5 &  19.8 &    21.0 &  17.6 \\
MoE Dom.sp. + DR &      30.7 &  30.2 &  24.9 &    28.1 &  28.1 \\
Tr. x5 + tags      &      30.8 &  29.3 &  25.6 &    25.8 &  26.7\\
Tr. x5 + tags + DR &      31.2 &  30.5 &  29.5 &    30.1 &  30.2 \\
\bottomrule
\end{tabular}
\end{small}
    \caption{BLEU scores for the "Subtitles" test set, when decoding with various domain labels listed in column names.}
    \label{tab:wrong_label_bleu_deen}
\end{table}

\paragraph{Robustness to wrong domain labels.} One of the important problems in multi-domain translation in the setting with access to domain labels, is that in practice domain labels can be erroneous, and models are very sensitive to these errors~\cite{pham-etal-2021-revisiting}. Table~\ref{tab:wrong_label_bleu_deen} shows substantial performance drops in BLEU scores on the "Subtitles" test set, when it is decoded with wrong domain labels by models trained without DR. Surprisingly, we find that the proposed domain randomization substantially reduces performance drops caused by wrong domain labels, possibly due to decreased overfitting to domain labels. 
Appendix~\ref{app:wrong_labels} confirms the same effect for other seen domains.

\begin{figure}
    \centering
    \includegraphics[width=\columnwidth]{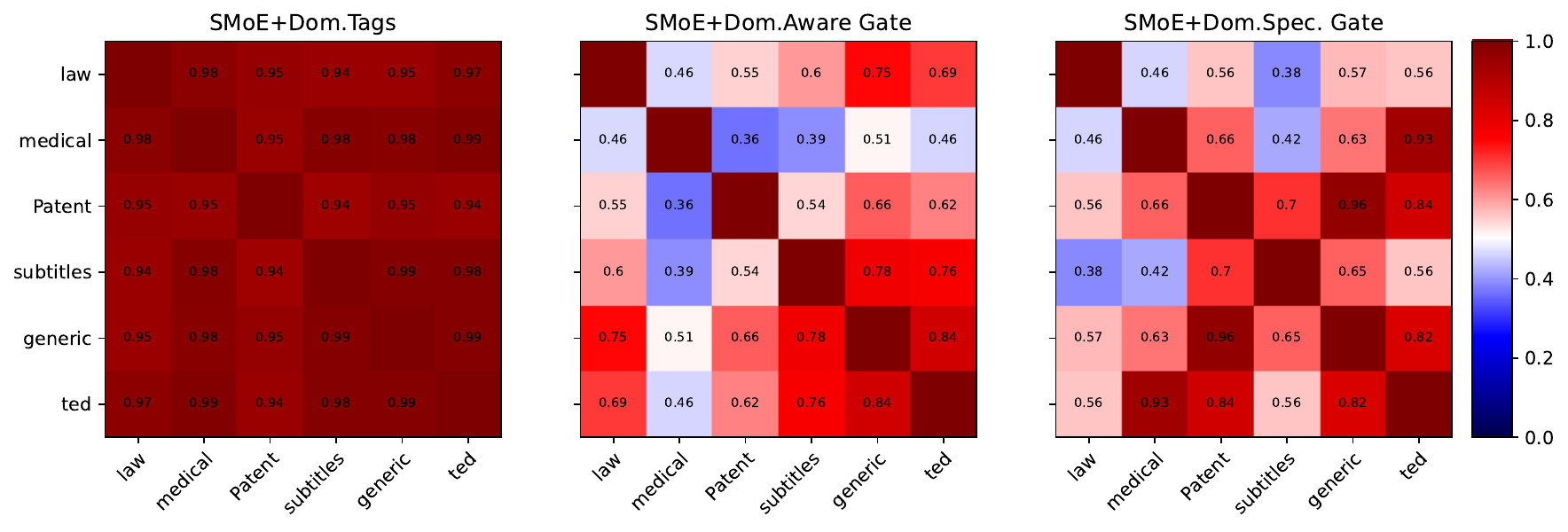}
    \includegraphics[width=\columnwidth]{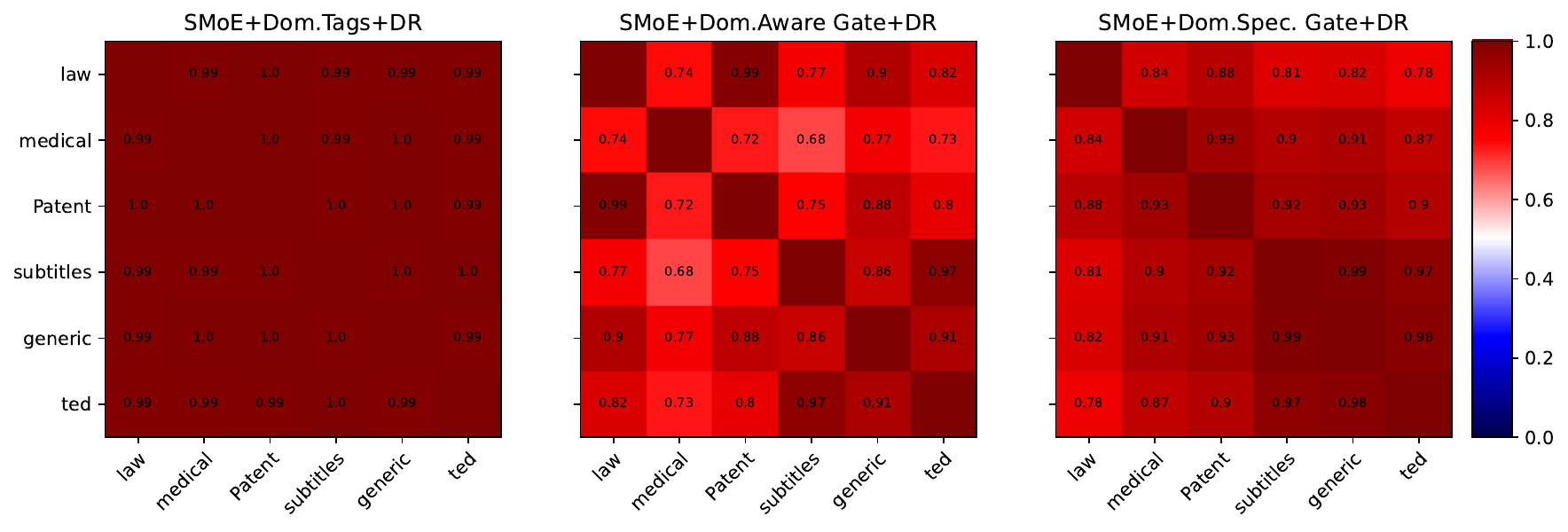}
    
    \caption{Pairwise similarity of experts activated with different domain embeddings.}
    \label{fig:experts_spec_by_dom_tag}
\end{figure}

\paragraph{SMoE analysis.} Figure \ref{fig:experts_spec_by_dom_tag} compares experts activated with different domain labels for different models. We see that SMoE+Dom.aware gate and SMoE+Dom.Spec.gate rely more on domain labels compared to SMoE+Dom.Tags, as well as models with DR. This explains higher performance drop when decoding with wrong labels. Domain randomization makes reduces theses models reliance on domain labels thus making them more robust to wrong labels. More in-depth anaylsis of gate statics is reported in Appendix \ref{app:gate_stats}.

\begin{table}[]
    \centering
    \begin{small}
    \begin{tabular}{p{4cm}|p{1cm}p{1cm}}
    \toprule
      & Seen  & Unseen  \\ \midrule
     \multicolumn{3}{c}{Transformer Base} \\ \midrule
     Seen domains only &  40.6  & 21.2  \\
     Generic data only &  34.1  &  26.7 \\
     Seen (50\%) + Gen. (50\%) &  40.3  & 26.5  \\
     Seen + Gen. (natural \%) &  38.3  &  26.7 \\ \midrule
     \multicolumn{3}{c}{Seen domains (50\%) + Generic data (50\%)} \\\midrule
     Transformer width $\times$5 &  41.9  & 26.9  \\
     Mixture-of-Experts &  41.7  &  27.0 \\
     \bottomrule
    \end{tabular}
    \end{small}
    \caption{BLEU scores, averages over seen and over unseen domains (excl. ``Pat (med)'' and ``Subt (med)''), for De-En translation, for the Transformer Base model trained on various data mixtures, and for two scaled models trained on the mixture of seen domains and the downsampled generic data.}
    \label{tab:exp_data_proportions}
\end{table}

\paragraph{Necessity of generic data and model scaling for multi-domain models.}
\label{sec:res:1}
Table~\ref{tab:exp_data_proportions} shows the average performance on seen and unseen domains (De-En), of Transformer Base (TB) models trained with different data mixtures. We observe that training solely on domain data leads to high performance on seen domains but very low performance on unseen domains. Including the generic corpora in the training data mixture substantially increases performance on unseen domains. However, generic corpora \textit{downsampling} is required to keep relatively high performance on seen domains, which is still lower than that of the model trained solely on the seen domain data.\footnote{We use the downsampling scheme, 50\% generic + 50\% seen domains, to train all the models in the main experiments.} The highest performance on both seen and unseen domains is achieved by two larger models, dense Transformer with the increased width and SMoE.

\section{Conclusion} 
In this work we investigate the potential of the SMoE architecture for effective, efficient and robust multi-domain translation. Our experiments lead us to generally 
%negative 
non-conclusive results regarding the utility of SMoE but highlight the necessity of model scaling and advantages of the straightforward width scaling. The width-scaled model performs on par with SMoE and adds almost no overhead at inference on modern GPUs compared to the non-scaled model. At the same time, for older and less optimized GPUs, SMoE may be a more efficient scaling approach. We also introduce domain randomization, a technique well suitable for scaled models which improves robustness on unseen domains, while preserving same performance on seen domains. We also find that DR improves robustness to wrong domain labels. 

\section{Limitations and discussion}
This work is limited to the setting where all the seen domains are available at training time, i.e. we do not consider  adaptation to newly appearing domains scenario after the model is trained. We leave this aspect for the future work. 

Another limitation is that the notion of domain we rely on is restricted to the origin of the data. These labels can be quite fuzzy and noisy as pointed out by \citep{saunders2022}: some datasets can share same lexicon choice, while others could share similar \textit{genre}.  Our gate activation analysis confirms that the notion of the domain is fuzzy, and the most effective and robust model (SMoE+Tags+DR) relies more on the input tokens rather than on domain labels. Nevertheless, it is still beneficial to have access to this knowledge (both at training and inference time) especially for the domains which are less represented in the training data. 

Finally, all our analysis is performed on Transformer.base architecture which is further scaled either via wider FF layers, or via SMoE. We believe this is quite common setting for most of the researchers who could benefit from our findings. We expect that much larger baseline (eg. comparable to Tr x5) may not benefit from scaling to the same extent when trained on small set of domains. We are quite confident though that, similar to the findings in multilingual NMT,  the need for scaling model size would persist as long as we need to accommodate more domains within the same model.

\bibliography{anthology,custom}
\bibliographystyle{ieeenat_fullname}

\appendix
\clearpage

\section{Experimental setup}
\label{app:exp_setup}

\begin{table}[ht!]
\normalsize
\centering
%\begin{tabular}{cp{1cm}p{1cm}p{1cm}p{1cm}p{1cm}p{1cm}}
\begin{tabular}{l|p{1cm}p{1cm}p{1cm}p{1cm}}
\toprule
\textbf{Domain} & \textbf{Train (M)} & \textbf{Valid (K)} & \textbf{Test (K)}  & \textbf{Sampl. prob.} \\ \midrule
%\multicolumn{5}{c}{Seen domains} \\ \midrule
Generic & 33.8 & 0.5 & n/a & 50\%\\
Europarl & 1.70 & 0.5 & n/a & 18\%\\
Patents & 1.00 & 2.0 & 2.0 & 10\%\\ 
Subtitles & 0.50 & 2.0 & 2.0 & 5.3\% \\
Law & 0.47 & 2.0 & 2.0 & 5.0\% \\
News-com. & 0.35 & 0.5 & n/a & 3.7\%\\
Medical & 0.25 & 2.0 & 2.0 & 2.6\%\\
Ted & 0.17 & 4.0 & 4.0 & 1.8\%\\ %\midrule 
Wikititles & 0.17 & 0.5 & n/a & 1.8\%\\
%\multicolumn{5}{c}{Unseen domains} \\ \midrule
%Flores & & 1K \\
%IT & & & 2K \\
%Koran & & & 2K \\
%Med. abst. & & 2K\\
%Pat. (med.) & & 2K \\
%Subt. (med.) & & 2K \\
\bottomrule
\end{tabular}
\caption{English-German data statistics (line pairs) for seen domains. Sampl. prob. denotes the probability of sampling line pairs from the domain in the training data mixture.}
\label{tab:data_de}
\end{table}

\begin{table}[ht!]
\normalsize
\centering
%\begin{tabular}{cp{1cm}p{1cm}p{1cm}p{1cm}p{1cm}p{1cm}}
\begin{tabular}{l|p{1cm}p{1cm}p{1cm}p{1cm}}
\toprule
\textbf{Domain} & \textbf{Train (M)} & \textbf{Valid (K)} & \textbf{Test (K)}  & \textbf{Sampl. prob.} \\ \midrule
%\multicolumn{5}{c}{Seen domains} \\ \midrule
Generic & 51.0  & 0.5  & n/a & 50.0\%\\
Europarl & 1.00  & 0.5  & n/a & 11.7\% \\
Patents & 1.00  & 2.0  & 2.0  & 11.7\%\\ 
Subtitles &  1.00 & 2.0  & 2.0  & 11.7\%\\ 
Law & 0.50  & 2.0  & 2.0  & 5.91\%\\ 
Medical & 0.40  & 2.0  & 2.0  & 4.76\%\\ 
Ted & 0.26  & 4.0  &  4.0  & 2.26\%\\ 
News-com.  & 0.15  & 0.5 & n/a & 1.79\% \\
\bottomrule
\end{tabular}
\caption{English-French data statistics (line pairs) for seen domains. Sampl. prob. denotes the probability of sampling the domain in the training data mixture.}
\label{tab:data_fr}
\end{table}

\paragraph{Data.}
We experiment with the German-English (De-En), English-German (En-De) and English-French (En-Fr) translation directions. 

All models are trained on a mix of generic data (Paracrawl, \citealp{paracrawl}) and several \textit{seen} domains: 
\begin{itemize}[noitemsep,topsep=0pt,parsep=0pt,partopsep=0pt]
\item Medical (European Medicines Agency documents, \citealp{emea});
\item Law (JRC-Acquis legislative EU texts, \citealp{acquis});
\item Ted (TED talks transcriptions, \citealp{qi-etal-2018-pre});
\item Subtitles (OpenSubtitles dataset, \citealp{opensubtitles16});
\item Patents (PatTR dataset, \citealp{PatTR});
\item Europarl (European Parliament Proceedings);
\item News-commentary;
\item Wikititles\footnote{\url{https://linguatools.org/tools/corpora/wikipedia-parallel-titles-corpora/}}.
\end{itemize}
The selection of seen domains is motivated by having  domains of different sizes and having a pair of close domains, TED Talks and Subtitles (they both relate to spoken language). We also test the generalization capabilities of models on a set of \textit{unseen} domains:
\begin{itemize}[noitemsep,topsep=0pt,parsep=0pt,partopsep=0pt]
\item IT (manuals and localization files of open-source software: GNOME, KDE, PHP, Ubuntu, and
OpenOffice);
\item Koran;
\item Flores (Wikipedia sentences, \citealp{flores});
\item Patents (medical);
\item Subtitles (medical);
\item Medical abstracts (MuchMore dataset\footnote{\url{https://muchmore.dfki.de/resources1.htm}, available only for En-De.});
\item European Central Bank documentation (ECB, \citealp{opus}, used only for En-Fr);
\item World bank data (used only for En-Fr).
\end{itemize}
Medical abstracts, Patents (medical), and Subtitles (medical)  are obtained from the UFAL Medical Corpus v. 1.0\footnote{\url{https://ufal.mff.cuni.cz/ufal_medical_corpus}}, and all the medical examples from this corpus are excluded from the training data.  The purpose of introducing Medical abstracts, Patents (medical), and Subtitles (medical) domains is to test models' capabilities to recognize unseen domains related to some seen domains (Medical, Patents, and Subtitles). Europarl, News-commentary, Medical, Law, Ted, Subtitles, IT, Koran, and ECB data is a part of the OPUS collection~\cite{opus}. %

Tables~\ref{tab:data_de} and~\ref{tab:data_fr} list data statistics. During training all the models except the experiment reported in Section~\ref{sec:res:1}, we sample the generic domain with probability 50\% and other domains with probabilities proportional to their size in terms of line pairs.

\paragraph{Training.} All the models are trained from scratch for 100k updates in mixed precision on four A100 or V100 GPUs, with a batch size of maximum 4000 tokens and accumulated gradients over 16 batches. We do not use early stopping, as usually all validation curves continue growing until the end of the training. Only data parallelism is used, model parallelism is not needed as all the models fit into a single GPU's memory. We use the Adam optimizer with the inverse square root learning rate schedule, with a warmup of 4k steps and a maximum learning rate of 0.001. The data is tokenized using Byte Pair Encoding with a vocabulary of size 24K for English-German and 16K for English-French.

\section{Additional results for the setting with wrong domain labels}
\label{app:wrong_labels}

\begin{table}[]
\centering
\begin{small}
\begin{tabular}{l|p{0.45cm}p{0.45cm}p{0.45cm}p{0.45cm}p{0.45cm}}
\toprule
{} &  Subt &    Ted &    Law &  Med &  Pat \\
\midrule
MoE + tags         &    50.41 &  44.44 &  39.76 &      42.08 &  41.38 \\
MoE + tags + DR    &    50.88 &  50.25 &  47.32 &      49.97 &  50.42 \\
MoE Dom.aw.        &    50.92 &  36.44 &  33.45 &      36.97 &  36.95 \\
MoE Dom.aw. + DR   &    50.77 &  49.74 &  41.96 &      41.73 &  49.42 \\
MoE Dom.spec.      &    50.94 &  36.34 &  33.45 &      34.47 &  34.85 \\
MoE Dom.spec. + DR &    50.29 &  48.02 &  46.07 &      47.31 &  47.17 \\
Tr. x5 + tags      &    50.55 &  46.01 &  37.86 &      41.73 &  44.49 \\
Tr. x5 + tags + DR &    50.41 &  50.06 &  48.15 &      49.55 &  50.02 \\
\bottomrule
\end{tabular}
\end{small}
    \caption{BLEU scores for the De-En "Medical" test set, when decoding with various domain labels listed in column names.}
    \label{tab:wrong_label_bleu_deen_med}
\end{table}

\begin{table}[]
\centering
\begin{small}
\begin{tabular}{l|p{0.45cm}p{0.45cm}p{0.45cm}p{0.45cm}p{0.45cm}}
\toprule
{} &  Subt &    Ted &    Law &  Med &  Pat \\
\midrule
MoE + tags         &  54.48 &    47.96 &  41.60 &      45.36 &  44.16 \\
MoE + tags + DR    &  54.29 &    54.00 &  52.11 &      53.49 &  54.04 \\
MoE Dom.aw.        &  54.98 &    36.74 &  36.48 &      38.41 &  32.64 \\
MoE Dom.aw. + DR   &  54.57 &    51.02 &  45.20 &      43.75 &  54.19 \\
MoE Dom.spec.      &  54.71 &    38.20 &  32.12 &      32.90 &  26.02 \\
MoE Dom.spec. + DR &  54.98 &    52.64 &  44.20 &      44.31 &  47.46 \\
Tr. x5 + tags      &  54.90 &    49.87 &  43.73 &      46.68 &  47.97 \\
Tr. x5 + tags + DR &  54.62 &    54.58 &  52.92 &      53.96 &  54.68 \\
\bottomrule
\end{tabular}
\end{small}
    \caption{BLEU scores for the De-En "Law" test set, when decoding with various domain labels listed in column names.}
    \label{tab:wrong_label_bleu_deen_law}
\end{table}

Tables~\ref{tab:wrong_label_bleu_deen_med} and~\ref{tab:wrong_label_bleu_deen_law} show performance on "Medical" and "Law" domains, when decoding with various domain labels. The tables confirm the conclusion made in the main text that using the proposed domain  randomisation improves model robustness to wrong domain labels at inference time.

%\section{Appendix}
%\label{sec:appendix}

%\subsection{English-German results}
%\begin{table*}[h!]
%\normalsize
%\centering
%\begin{tabular}{cp{1cm}p{1cm}p{1cm}p{1cm}p{1cm}p{1cm}}
%\begin{tabular}{p{4.9cm}c|ccccc|c}
%\toprule
%\textbf{Model~/~Domain} & \textbf{\#Params} & \textbf{Law} & \textbf{Med} &   \textbf{Ted} & \textbf{Subt} & \textbf{Pat} &  \textbf{Avg.} \\
%\midrule
%\multicolumn{9}{c}{Without access to seen domains labels} \\ \midrule
%Transformer Base (TB)  & 56M &   66.72& 63.85& 60.0& 48.42& 50.67& 57.93\\
%Mixture-of-Experts top-1   &  170M & &  &  & & & & \\
%Mixture-of-Experts (SMoE) &  170M  & 68.39& 64.93& 60.44& 49.39& 51.13 & 58.86\\
%SMoE & PARAMS & 68.39& 64.93& 60.44& 49.39& 51.13
%Transformer width x2  & 70M     &- & - & - & -& -  \\ \midrule

%\multicolumn{9}{c}{With access to seen domains labels} \\ \midrule
%\textit{TB + domain tags}     &  56M  & 66.74& 63.64& 60.27& 48.89& 50.68& 58.04
 %TODO: these results corresponds to 80k updates, will update it with full results asap

%SMoE + domain tags + DR  & 170M&  68.17& 64.98& 60.96& 49.21& 51.06 & 58.88\\
%SMoE + Dom.Aware Gate + DR  & 170M   & 68.15& 65.52& 60.83& 49.35& 51.07 & 58.98\\
%SMoE.tags & PARAMS & 68.17& 64.98& 60.96& 49.21& 51.06
%daSMoE_rep02 & PARAMS & 68.15& 65.52& 60.83& 49.35& 51.07

%\bottomrule
%\end{tabular}
%\caption{Results (chrF) for seen domains, English-German translation. DR: domain randomization.}
%\label{tab:main_seen_ende_chrf}
%\end{table*}

\begin{table*}[h!]
\normalsize
\centering
%\begin{tabular}{cp{1cm}p{1cm}p{1cm}p{1cm}p{1cm}p{1cm}}
\begin{tabular}{p{4.9cm}|ccccc|c}
\toprule
\textbf{Model}  & \textbf{Law} & \textbf{Med} &   \textbf{Ted} & \textbf{Subt} & \textbf{Pat} &  \textbf{Avg.} \\
\midrule
\multicolumn{7}{c}{Without access to seen domains labels} \\ \midrule
Transformer Base (TB)  &  \underline{45.6}  &  43.5  &  \underline{34.2}  &  25.0  &  \underline{28.4}  &  35.3\\
Transformer width x1.5$^*$  &  46.2  &  43.5  &  34.7  &  25.4  &  28.6  &  35.7\\
Transformer width x5  &  \textbf{48.2}  &  \textbf{44.3}  &  \textbf{35.2}  &  \textbf{26.2}  &  \textbf{29.0}  &  \textbf{36.6}\\
Mixture of Experts (MoE)  &  \textbf{48.0}  &  \textbf{44.6}  &  \textbf{35.2}  &  25.6  &  \textbf{28.9}  &  \textbf{36.5}\\
%Transformer Base (TB) &  45.62& 44.02& 34.27& 25.01& 28.38& 35.46\\
%Transformer Base (TB) & 45.66& 43.5& 34.2& 24.84& 28.41& 35.32\\
%Transformer width $\times$1.5 & 46.18& 44.06& 34.76& 25.54& 28.61& 35.83\\
%Transformer width $\times$1.5 & 46.24& 43.52& 34.71& 25.45& 28.66& 35.72\\
%Transformer width $\times$5 & 48.2& 44.89& 35.32& 26.31& 28.94& 36.73\\
%Transformer width $\times$5  & 48.18& 44.32& 35.22& 26.18& 28.99& 36.58\\
%Mixture-of-Experts (SMoE)  & 48.0& 45.16& 35.2& 25.76& 28.84& 36.59 \\
%SMoE 50-2 & 50.74& 47.72& 34.57& 26.21& 28.61& 37.57
%Mixture-of-Experts (SMoE) & 48.01& 44.58& 35.16& 25.65& 28.85& 36.45\\
%SMoE & PARAMS & 68.39& 64.93& 60.44& 49.39& 51.13
%Transformer width x2   &  - & -& - & - & -& -  
\midrule
\multicolumn{7}{c}{With access to seen domains labels} \\ \midrule
%\textit{TB + domain tags}     & 45.59& 43.45& 34.9& 26.55& 28.35& 35.77 \\ 
%SMoE + domain tags + DR &  47.7& 45.57& 35.65& 26.41& 28.66& 36.8 \\
%SMoE + Dom.Aware Gate  &  47.66& 46.74& 35.66& 26.68& 28.78& 37.1 \\
%SMoE + Dom.Aware Gate + DR  & 47.57& 45.93& 35.8& 26.64& 28.77& 36.94\\
TB + domain tags  &  \underline{45.6}  &  \underline{42.9}  &  \underline{34.8}  &  26.5  &  28.4  &  35.6\\
Tr. x1.5 + domain tags$^*$  &  46.2  &  43.8  &  35.3  &  26.3  &  28.5  &  36.0\\
Tr. x5 + domain tags + DR  &  48.5  &  44.8  &  36.1  &  26.5  &  28.9  &  \textbf{36.9}\\
TB + domain adapters  &  \textbf{48.2}  &  \textbf{46.0}  &  \underline{33.5}  &  \underline{25.3}  &  28.7  &  36.3\\
MoE + domain tags  &  \textbf{48.2}  &  \textbf{45.1}  &  \textbf{35.8}  &  26.4  &  28.5  &  \textbf{36.8}\\
MoE + domain tags + DR  &  \textbf{47.6}  &  \textbf{45.1}  &  \textbf{35.6}  &  26.3  &  28.7  &  \textbf{36.7}\\
MoE + Dom. Aware Gate  &  \textbf{47.8}  &  \textbf{46.1}  &  35.6  &  26.5  &  \textbf{28.8}  &  \textbf{37.0}\\
MoE + Dom. Aware Gate + DR  &  \textbf{47.8}  &  \textbf{44.6}  &  35.5  &  26.4  &  28.7  &  \textbf{36.6}\\
%\textit{TB + domain tags} & 45.59& 42.92& 34.77& 26.47& 28.37& 35.62\\
%TB $\times$1.5 + domain tags & 46.24& 43.78& 35.33& 26.26& 28.54& 36.03\\
%TB + domain adapters & 48.22& 45.98& 33.5& 25.31& 28.7& 36.34\\
%SMoE + domain tags & 48.19& 45.06& 35.77& 26.41& 28.53& 36.79\\
%SMoE + domain tags + DR & 47.62& 45.06& 35.6& 26.33& 28.69& 36.66\\
%SMoE + Dom.Aware Gate & 47.79& 46.1& 35.64& 26.53& 28.82& 36.98\\
%SMoE + Dom.Aware Gate + DR & 47.85& 44.6& 35.52& 26.41& 28.66& 36.61\\
%SMoE.tags & PARAMS & 68.17& 64.98& 60.96& 49.21& 51.06
%daSMoE_rep02 & PARAMS & 68.15& 65.52& 60.83& 49.35& 51.07
\bottomrule
\end{tabular}
\caption{Results (BLEU) for seen domains, English-German translation. DR: domain randomization.  \textbf{Bold} denotes significant improvement vs. "Transformer width $\times$1.5" (without domain labels) or "Tr. $\times$1.5 + domain tags" (with domain labels), both marked with $^*$; \underline{underline} denotes significant losses.}
\label{tab:main_seen_ende_bleu}
\end{table*}

%\begin{table*}[h!]
%\normalsize
%\centering
%\begin{tabular}{cp{1cm}p{1cm}p{1cm}p{1cm}p{1cm}p{1cm}}
%\begin{tabular}{p{4.25cm}|p{1cm}p{1cm}p{1cm}p{1.5cm}p{1.3cm}p{1.3cm}|c}
%\toprule
%\textbf{Model~/~Domain}   & \textbf{It} & \textbf{Koran} & \textbf{Wiki} & \textbf{Medical abstracts} & \textbf{Patents (med.)} & \textbf{Subtitles (med.)} &  \textbf{Avg.} \\
%\midrule
%\multicolumn{8}{c}{Without access to seen domains labels} \\ \midrule
%Transformer Base (TB)   &  \\
%Mixture-of-Experts (SMoE)   & 64.25& 51.49& 36.34& 41.29& 54.92& 55.6& 51.68 \\
%Transformer width x2  &  &  & & & &\\
%\multicolumn{8}{c}{With access to seen domains labels} \\ \midrule
%TB + domain tags     &   \\
%SMoE + domain tags + DR   &  64.1& 51.66& 36.41& 41.28& 54.97& 55.64& 51.69 \\
%SMoE + Dom.Aware Gate + DR  & 64.05& 51.66& 36.58& 41.09& 54.94& 55.38& 51.61 \\
%\bottomrule
%\end{tabular}
%\caption{Results for unseen domains (chrF), English-German translation. For models with access to seen domain labels (second group of rows), the ``generic'' label is used for unseen domains. DR: domain randomization.}
%\label{tab:main_unseen_ende}
%\end{table*}

\begin{table*}[h!]
\normalsize
\centering
\begin{tabular}{p{5.5cm}|cccp{1cm}p{1cm}p{1cm}|c}
\toprule
\textbf{Model}   & \textbf{Flores} & \textbf{IT} & \textbf{Koran} & \textbf{Pat (med)} & \textbf{Subt (med)} & \textbf{Med abst} &  \textbf{Avg} \\
\midrule
\multicolumn{8}{c}{Without access to seen domains labels} \\ \midrule
Transformer Base (TB)  &  \underline{36.4}  &  28.1  &  10.8  &  \underline{31.9}  &  32.9  &  11.1  &  25.2\\
Transformer width x1.5$^*$  &  37.1  &  28.5  &  10.9  &  32.2  &  33.3  &  11.2  &  25.5\\
Transformer width x5  &  \textbf{37.8}  &  \textbf{29.2}  &  \textbf{11.6}  &  32.2  &  33.7  &  \textbf{11.5}  &\textbf{  26.0}\\
Mixture of Experts (MoE)  &  36.8  &  28.4  &  11.0  &  32.3  &  33.8  &  11.2  &  25.6\\
%Transformer Base (TB)   & 36.38& 28.17& 11.05& 11.1& 31.86& 33.01& 25.65\\
%Mixture-of-Experts (SMoE)   & 36.8& 28.55& 11.34& 11.24& 32.3& 34.0& 26.1\\
%Transformer width $\times$1.5  &37.06& 28.68& 11.2& 11.23& 32.17& 33.41& 26.02\\ 
%Transformer width $\times$5 & 37.84& 29.35& 11.92& 11.47& 32.2& 33.84& 26.53\\
%SMoE 50-2 &  36.86& 28.79& 12.46& 11.44& 31.56& 33.4& 26.17\\
\midrule
\multicolumn{8}{c}{Without access to seen domains labels} \\ \midrule
TB + domain tags  &  \underline{35.7}  &  28.2  &  10.9  &  30.7  &  31.0  &  11.2  &  24.6\\
Tr. x1.5 + domain tags$^*$  &  36.3  &  28.4  &  10.8  &  30.9  &  31.4  &  11.1  &  24.8\\
Tr. x5 + domain tags + DR  &  37.8  &  28.6  &  11.5  &  32.3  &  34.2  &  11.4  &  \textbf{26.0}\\
TB + domain adapters  &  \underline{35.0}  &  28.8  &  10.8  &  \underline{23.5}  &  \underline{29.6}  &  \underline{10.8}  &  \underline{23.1}\\
MoE + domain tags  &  36.3  &  28.6  &  \textbf{11.6}  &  30.8  &  31.3  &  \textbf{11.4}  &  25.0\\
MoE + domain tags + DR  &  36.5  &  28.5  &  \textbf{11.2}  &  \textbf{32.4}  &  \textbf{33.7}  &  11.2  &  \textbf{25.6}\\
MoE + Dom. Aware Gate  &  36.2  &  \textbf{29.1}  &  \textbf{11.3}  &  \underline{27.6}  &  31.3  &  11.2  &  24.4\\
MoE + Dom. Aware Gate + DR  &  \textbf{37.5}  &  28.5  &  \textbf{11.3}  &  \textbf{32.1}  &  \textbf{33.1}  &  \textbf{11.3}  &  \textbf{25.6}\\Transformer Base (TB)      &      36.4 &  28.1 &  10.8 &              31.9 &                32.9 &              11.1 &  25.2 \\
%TB + domain tags     &  35.68& 28.38& 11.3& 11.15& 30.75& 31.23& 25.16 \\
%SMoE + domain tags + DR   &36.46& 28.76& 11.5& 11.22& 32.44& 33.86& 26.06 \\
%SMoE + Dom.Aware Gate & 36.22& 29.24& 11.58& 11.2& 27.57& 31.48& 25.06\\
%SMoE + Dom.Aware Gate + DR  & 36.74& 28.75& 11.92& 10.99& 32.29& 33.55& 26.04\\
\bottomrule
\end{tabular}
\caption{Results for unseen domains (BLEU), English-German translation. For models with access to seen domain labels (second group of rows), the ``generic'' label is used for unseen domains. DR: domain randomization.  \textbf{Bold} denotes significant improvement vs. "Transformer width $\times$1.5" (without domain labels) or "Tr. $\times$1.5 + domain tags" (with domain labels), both marked with $^*$; \underline{underline} denotes significant losses.}
\label{tab:main_unseen_ende_bleu}
\end{table*}
%\subsection{English-French results}

\begin{table*}[h!]
\normalsize
\centering
%\begin{tabular}{cp{1cm}p{1cm}p{1cm}p{1cm}p{1cm}p{1cm}}
\begin{tabular}{p{4.9cm}|ccccc|c}
\toprule
\textbf{Model} & \textbf{Law} & \textbf{Med} &   \textbf{Ted} & \textbf{Subt} & \textbf{Pat} &  \textbf{Avg} \\
\midrule
\multicolumn{7}{c}{Without access to seen domains labels} \\ \midrule
Transformer Base (TB)  &  \underline{59.7}  &  53.7  &  44.8  &  \underline{32.0}  &  \underline{37.0}  &  \underline{45.5}\\
Transformer width $\times$1.5$^*$  &  60.6  &  53.7  &  45.1  &  32.6  &  37.4  &  45.9\\
Transformer width $\times$5  &  \textbf{62.6}  &  \textbf{56.2}  &  \textbf{45.5}  &  \textbf{33.4}  &  37.5  &  \textbf{47.0}\\
Mixture of Experts (SMoE)  &  \textbf{62.5}  &  \textbf{56.3}  &  45.4  &  32.9  &  37.3  &  \textbf{46.9}\\
\midrule
\multicolumn{7}{c}{With access to seen domains labels} \\ \midrule
TB + domain tags  &  \underline{59.8}  &  \underline{53.3}  &  44.9  &  32.6  &  \underline{37.0}  &  \underline{45.5}\\
TB + domain tags + DR  &  \underline{60.1}  &  \underline{53.5}  &  44.8  &  32.5  &  \underline{37.1}  &  \underline{45.6}\\
Tr. $\times$1.5 + domain tags$^*$  &  61.2  &  54.5  &  44.9  &  32.7  &  37.3  &  46.1\\
Tr. 5 + domain tags + DR  &  \textbf{62.5}  &  \textbf{56.0}  &  \textbf{45.3}  &  \textbf{33.4}  &  37.5  &  \textbf{46.9}\\
TB + domain adapters  &  \textbf{62.5}  &  55.2  &  \underline{42.0}  &  \underline{31.7}  &  37.1  &  45.7\\
TB + domain adapters + DR  &  \underline{60.4}  &  54.2  &  \underline{43.2}  &  \underline{31.6}  &  \underline{36.8}  &  \underline{45.2}\\
SMoE + domain tags  &  \textbf{62.9}  &  \textbf{56.7}  &  \textbf{45.3}  &  33.1  &  \textbf{37.5}  &  \textbf{47.1}\\
SMoE + domain tags + DR  &  \textbf{62.8}  &  \textbf{56.9}  &  45.2  &  \textbf{33.3}  &  37.4  &  \textbf{47.1}\\
SMoE + Dom. Aware Gate  &  \textbf{62.9}  &  \textbf{56.8}  &  \underline{44.4}  &  32.6  &  37.3  &  \textbf{46.8}\\
SMoE + Dom. Aware Gate + DR  &  \textbf{62.3}  &  \textbf{56.8}  &  45.2  &  33.0  &  37.5  &  \textbf{47.0}\\
SMoE + Dom. Spec. Gate  &  \textbf{62.5}  &  \textbf{56.3}  &  45.2  &  33.2  &  37.3  &  \textbf{46.9}\\
SMoE + Dom. Spec. Gate + DR  &  62.8  &  56.0  &  45.5  &  33.2  &  37.6  &  \textbf{47.0}\\
\bottomrule
\end{tabular}
\caption{BLEU scores for seen domains, English-French translation. DR: domain randomization.  \textbf{Bold} denotes significant improvement vs. "Transformer width $\times$1.5" (without domain labels) or "Tr. $\times$1.5 + domain tags" (with domain labels), both marked with $^*$; \underline{underline} denotes significant losses.}
\label{tab:enfr_seen_bleu}
\end{table*}

% BLEU ENFR UNSEEN
\begin{table*}[ht!]
\normalsize
\centering
%\begin{tabular}{cp{1cm}p{1cm}p{1cm}p{1cm}p{1cm}p{1cm}}
\begin{tabular}{p{4.6cm}|cccp{1cm}p{1cm}p{1cm}c|c}
\toprule
\textbf{Model}   & \textbf{Flores} & \textbf{IT} & \textbf{Koran} & \textbf{Pat (med)} & \textbf{Subt (med)}& \textbf{World bank} & \textbf{ECB}  &  \textbf{Avg} \\
\midrule
\multicolumn{7}{c}{Without access to seen domains labels} \\ \midrule
Transformer Base (TB)  &  47.5  &  \underline{16.0}  &  \underline{12.0}  &  \underline{35.5}  &  \underline{32.0}  &  32.4  &  44.2  &  31.3\\
Transformer width $\times$1.5$^*$  &  47.5  &  16.2  &  12.8  &  35.8  &  32.7  &  32.4  &  44.8  &  31.7\\
Transformer width $\times$5  &  \textbf{48.4}  &  16.4  &  \textbf{13.8}  &  36.0  &  32.9  &  \textbf{32.9}  &  44.9  &  \textbf{32.2}\\
Mixture of Experts (SMoE)  &  47.8  &  16.4  &  13.2  &  35.8  &  33.0  &  \textbf{32.8}  &  44.7  &  32.0\\
\midrule
\multicolumn{7}{c}{With access to seen domains labels} \\ \midrule
TB + domain tags  &  46.9  &  16.0  &  \underline{12.0}  &  34.3  &  29.4  &  32.0  &  \underline{43.4}  &  30.6\\
TB + domain tags + DR  &  47.2  &  16.0  &  12.3  &  \textbf{35.6}  &  \textbf{32.5}  &  \textbf{32.3}  &  44.6  &  \textbf{31.5}\\
Tr. $\times$1.5 + domain tags$^*$  &  46.5  &  16.2  &  12.7  &  34.1  &  29.7  &  31.8  &  44.2  &  30.7\\
Tr. 5 + domain tags + DR  &  \textbf{48.5}  &  16.2  &  \textbf{13.9}  &  \textbf{36.1}  &  \textbf{33.2}  &  \textbf{32.7}  &  \textbf{45.2}  &  \textbf{32.3}\\
TB + domain adapters  &  46.7  &  16.1  &  12.9  &  \underline{29.5}  &  \underline{27.6}  &  31.9  &  \underline{41.8}  &  \underline{29.5}\\
TB + domain adapters + DR  &  \textbf{48.3}  &  15.9  &  12.3  &  \textbf{35.8}  &  \textbf{32.7}  &  \textbf{32.6}  &  \textbf{44.9}  &  \textbf{31.8}\\
SMoE + domain tags  &  46.9  &  16.4  &  \textbf{13.5}  &  \textbf{34.4}  &  29.5  &  32.1  &  44.4  &  31.0\\
SMoE + domain tags + DR  &  \textbf{48.3}  &  16.3  &  \textbf{13.4}  &  \textbf{35.9}  &  \textbf{33.2}  &  \textbf{33.0}  &  \textbf{45.2}  &  \textbf{32.2}\\
SMoE + Dom. Aware Gate  &  \textbf{47.6}  &  16.4  &  \textbf{13.3}  &  \underline{32.9}  &  29.4  &  32.1  &  \underline{43.4}  &  30.7\\
SMoE + Dom. Aware Gate + DR  &  \textbf{47.9}  &  16.3  &  12.9  &  \textbf{35.8}  &  \textbf{33.0}  &  \textbf{32.7}  &  \textbf{45.0}  &  \textbf{31.9}\\
SMoE + Dom. Spec. Gate  &  \textbf{48.1}  &  16.4  &  \textbf{13.3}  &  \underline{32.0}  &  29.2  &  \textbf{32.6}  &  44.1  &  30.8\\
SMoE + Dom. Spec. Gate + DR  &  48.5  &  16.3  &  12.8  &  35.9  &  32.9  &  32.7  &  44.9  &  \textbf{32.0}\\
\bottomrule
\end{tabular}
\caption{BLEU scores for unseen domains, English-French translation. For models with access to seen domain labels (second group of rows), the ``generic'' label is used for unseen domains. DR: domain randomization.  \textbf{Bold} denotes significant improvement vs. "Transformer width $\times$1.5" (without domain labels) or "Tr. $\times$1.5 + domain tags" (with domain labels), both marked with $^*$; \underline{underline} denotes significant losses.}
\label{tab:enfr_unseen_bleu}
\end{table*}

% COMET DEEN SEEN
\begin{table*}[h!]
\normalsize
\centering
%\begin{tabular}{cp{1cm}p{1cm}p{1cm}p{1cm}p{1cm}p{1cm}}
\begin{tabular}{p{4.9cm}|ccccc|c}
\toprule
\textbf{Model} & \textbf{Law} & \textbf{Med} &   \textbf{Ted} & \textbf{Subt} & \textbf{Pat} &  \textbf{Avg} \\
\midrule
\multicolumn{7}{c}{Without access to seen domains labels} \\ \midrule
Transformer Base (TB)  &  87.0  &  85.7  &  86.3  &  79.5  &  80.0  &  83.7\\
Transformer width $\times$1.5$^*$  &  87.1  &  85.6  &  86.4  &  79.7  &  80.0  &  83.8\\
Transformer width $\times$5  &  \textbf{87.5}  &  85.8  &  \textbf{86.6}  &  \textbf{80.1}  &  \textbf{80.3}  &  \textbf{84.1}\\
Mixture of Experts (SMoE)  &  \textbf{87.5}  &  \textbf{86.0}  &  \textbf{86.6}  &  \textbf{80.1}  &  \textbf{80.3}  &  \textbf{84.1}\\
\midrule
\multicolumn{7}{c}{With access to seen domains labels} \\ \midrule
TB + domain tags  &  \underline{87.1}  &  85.6  &  \underline{86.4}  &  79.4  &  \underline{80.0}  &  83.7\\
TB + domain tags + DR  &  \underline{87.1}  &  85.5  &  \underline{86.5}  &  79.7  &  80.1  &  83.8\\
Tr. $\times$1.5 + domain tags$^*$ &  87.3  &  85.7  &  86.6  &  79.6  &  80.2  &  83.8\\
TB + domain adapters  &  87.2  &  85.5  &  \underline{84.3}  &  \underline{77.9}  &  80.2  &  \underline{83.0}\\
TB + domain adapters + DR  &  87.1  &  85.4  &  \underline{85.6}  &  \underline{78.9}  &  80.1  &  \underline{83.4}\\
SMoE + domain tags  &  \textbf{87.6}  &  85.8  &  \textbf{86.8}  &  \textbf{80.1}  &  80.2  &  \textbf{84.1}\\
SMoE + domain tags + DR  &  \textbf{87.5}  &  \textbf{86.0}  &  \textbf{86.9}  &  \textbf{80.1}  &  80.2  &  \textbf{84.1}\\
SMoE + Dom. Aware Gate  &  \textbf{87.5}  &  \textbf{85.9}  &  86.7  &  79.8  &  80.2  &  \textbf{84.0}\\
SMoE + Dom. Aware Gate + DR  &  \textbf{87.5}  &  \textbf{86.0}  &  \textbf{86.8}  &  \textbf{80.1}  &  80.1  &  \textbf{84.1}\\
SMoE + Dom. Spec. Gate + DR  &  \textbf{87.5}  &  \textbf{85.9}  &  \textbf{86.8}  &  \textbf{80.1}  &  80.2  & \textbf{84.1}\\
\bottomrule
\end{tabular}
\caption{COMET results for seen domains, German-English translation. DR: domain randomization.  \textbf{Bold} denotes significant improvement vs. "Transformer width $\times$1.5" (without domain labels) or "Tr. $\times$1.5 + domain tags" (with domain labels), both marked with $^*$; \underline{underline} denotes significant losses.}
\label{tab:deen_seen_comet}
\end{table*}

% COMET DEEN UNSEEN
\begin{table*}[ht!]
\normalsize
\centering
%\begin{tabular}{cp{1cm}p{1cm}p{1cm}p{1cm}p{1cm}p{1cm}}
\begin{tabular}{p{4.6cm}|cccp{1cm}p{1cm}p{1cm}|c}
\toprule
\textbf{Model}   & \textbf{Flores} & \textbf{IT} & \textbf{Koran} & \textbf{Pat (med)} & \textbf{Subt (med)}& \textbf{Med abstr} &  \textbf{Avg} \\
\midrule
\multicolumn{7}{c}{Without access to seen domains labels} \\ \midrule
Transformer Base (TB)  &  88.4  &  82.4  &  70.8  &  82.3  &  82.0  &  70.1  &  79.3\\
Transformer width $\times$1.5$^*$  &  88.5  &  82.6  &  70.9  &  82.3  &  82.1  &  70.0  &  79.4\\
Transformer width $\times$5  &  \textbf{88.8}  &  82.7  &  70.7  &  \textbf{82.6}  &  \textbf{82.6}  &  \textbf{71.2}  &  \textbf{79.7}\\
Mixture of Experts (SMoE)  &  \textbf{88.8}  &  82.5  &  \textbf{71.3}  &  82.4  &  \textbf{82.3}  &  \textbf{70.4}  &  \textbf{79.6}\\
\midrule
\multicolumn{7}{c}{With access to seen domains labels} \\ \midrule
TB + domain tags  &  88.3  &  82.3  &  \underline{70.0}  &  \underline{82.0}  &  81.5  &  \underline{70.1}  &  \underline{79.0}\\
TB + domain tags + DR  &  88.3  &  82.2  &  \underline{69.8}  &  \textbf{82.4}  &  \textbf{81.9}  &  \underline{70.2}  &  79.1\\
Tr. $\times$1.5 + domain tags$^*$  &  88.5  &  82.5  &  70.5  &  82.2  &  81.6  &  70.6  &  79.3\\
TB + domain adapters  &  \underline{88.3}  &  82.4  &  \underline{70.2}  &  \underline{81.2}  &  \underline{80.6}  &  \underline{70.2}  &  \underline{78.8}\\
TB + domain adapters + DR  &  88.6  &  82.6  &  70.3  &  \textbf{82.4}  &  \textbf{82.1}  &  70.7  &  79.4\\
SMoE + domain tags  &  \textbf{88.8}  &  \textbf{83.0}  &  \textbf{71.4}  &  \textbf{82.3}  &  81.7  &  70.5  &  \textbf{79.6}\\
SMoE + domain tags + DR  &  \textbf{88.8}  &  82.3  &  \textbf{71.4}  &  \textbf{82.4}  &  \textbf{82.4}  &  70.5  &  \textbf{79.6}\\
SMoE + Dom. Aware Gate  &  \textbf{88.7}  &  \textbf{82.8}  &  \textbf{71.4}  &  \underline{81.1}  &  81.5  &  \underline{70.1}  &  79.3\\
SMoE + Dom. Aware Gate + DR  &  \textbf{88.8}  &  82.7  &  \textbf{71.2}  &  82.3  &  \textbf{82.2}  &  \underline{70.0}  &  \textbf{79.5}\\
SMoE + Dom. Spec. Gate + DR  &  \textbf{88.7}  &  82.4  &  \textbf{71.1}  &  \textbf{82.4}  &  \textbf{82.5}  &  \underline{70.4}  &  \textbf{79.6}\\
\bottomrule
\end{tabular}
\caption{COMET results for unseen domains, German-English translation. For models with access to seen domain labels (second group of rows), the ``generic'' label is used for unseen domains. DR: domain randomization.  \textbf{Bold} denotes significant improvement vs. "Transformer width $\times$1.5" (without domain labels) or "Tr. $\times$1.5 + domain tags" (with domain labels), both marked with $^*$; \underline{underline} denotes significant losses.}
\label{tab:deen_unseen_comet}
\end{table*}

% COMET ENDE SEEN
\begin{table*}[h!]
\normalsize
\centering
%\begin{tabular}{cp{1cm}p{1cm}p{1cm}p{1cm}p{1cm}p{1cm}}
\begin{tabular}{p{4.9cm}|ccccc|c}
\toprule
\textbf{Model} & \textbf{Law} & \textbf{Med} &   \textbf{Ted} & \textbf{Subt} & \textbf{Pat} &  \textbf{Avg} \\
\midrule
\multicolumn{7}{c}{Without access to seen domains labels} \\ 
\midrule
Transformer Base (TB)      &  87.0 &    84.0 &  83.2 &      76.7 &    76.8 &  81.6 \\
Transformer width x1.5     &  87.2 &    84.3 &  83.5 &      77.6 &    77.0 &  81.9 \\
Transformer width x5       &  87.6 &    84.1 &  84.2 &      78.3 &    77.2 &  82.2 \\
Mixture of Experts (MoE)   &  87.4 &    84.3 &  83.6 &      77.7 &    77.0 &  82.0 \\
\multicolumn{7}{c}{With access to seen domains labels} \\ 
\midrule
TB + domain tags           &  87.0 &    83.9 &  83.5 &      77.2 &    76.7 &  81.7 \\
Tr. x1.5 + domain tags     &  87.4 &    84.3 &  84.0 &      77.6 &    77.0 &  82.1 \\
TB + domain adapters       &  87.2 &    84.2 &  82.5 &      76.9 &    76.9 &  81.5 \\
MoE + domain tags          &  87.5 &    84.5 &  84.2 &      77.9 &    77.0 &  82.2 \\
MoE + domain tags + DR     &  87.3 &    84.4 &  84.1 &      78.0 &    77.1 &  82.2 \\
MoE + Dom. Aware Gate      &  87.4 &    84.6 &  84.1 &      77.9 &    77.1 &  82.2 \\
MoE + Dom. Aware Gate + DR &  87.5 &    84.3 &  84.2 &      77.8 &    77.0 &  82.2 \\
\bottomrule
\end{tabular}
\caption{COMET results for seen domains, English-German translation. DR: domain randomization.  \textbf{Bold} denotes significant improvement vs. "Transformer width $\times$1.5" (without domain labels) or "Tr. $\times$1.5 + domain tags" (with domain labels), both marked with $^*$; \underline{underline} denotes significant losses.}
\label{tab:ende_seen_comet}
\end{table*}

% COMET ENDE UNSEEN
\begin{table*}[ht!]
\normalsize
\centering
%\begin{tabular}{cp{1cm}p{1cm}p{1cm}p{1cm}p{1cm}p{1cm}}
\begin{tabular}{p{4.6cm}|cccp{1cm}p{1cm}p{1cm}|c}
\toprule
{} & Flores &    It & Koran & Pat (med) & Subtitles\_medical & Medical abstracts &  Avg. \\
\midrule
Transformer Base (TB)      &   84.3 &  78.1 &  69.2 &      78.7 &              79.0 &              67.7 &  76.2 \\
Transformer width x1.5     &   85.0 &  78.6 &  69.6 &      78.9 &              79.3 &              68.6 &  76.7 \\
Transformer width x5       &   85.6 &  78.9 &  70.4 &      79.2 &              79.7 &              69.5 &  77.2 \\
Mixture of Experts (MoE)   &   84.8 &  78.3 &  69.5 &      79.0 &              79.6 &              68.4 &  76.6 \\
TB + domain tags           &   83.9 &  79.0 &  69.1 &      78.0 &              77.0 &              69.2 &  76.1 \\
Tr. x1.5 + domain tags     &   84.5 &  79.1 &  69.2 &      78.0 &              77.6 &              68.1 &  76.1 \\
TB + domain adapters       &   83.0 &  79.1 &  69.0 &      71.7 &              76.2 &              67.1 &  74.3 \\
MoE + domain tags          &   84.0 &  79.2 &  69.7 &      78.0 &              77.3 &              68.3 &  76.1 \\
MoE + domain tags + DR     &   84.4 &  78.7 &  69.5 &      79.0 &              79.3 &              67.7 &  76.4 \\
MoE + Dom. Aware Gate      &   84.1 &  79.2 &  69.5 &      75.5 &              77.0 &              68.1 &  75.6 \\
MoE + Dom. Aware Gate + DR &   85.0 &  78.7 &  70.0 &      78.9 &              79.1 &              68.4 &  76.7 \\
\bottomrule

%\toprule
%\textbf{Model~/~Domain}   & \textbf{Flores} & \textbf{IT} & \textbf{Koran} & \textbf{Pat (med)} & \textbf{Subt (med)}& \textbf{Med abstr} &  \textbf{Avg} \\
%\midrule
%
\end{tabular}
\caption{COMET results for unseen domains, English-German translation, shallow decoder (3 layers). For models with access to seen domain labels (second group of rows), the ``generic'' label is used for unseen domains. DR: domain randomization.  \textbf{Bold} denotes significant improvement vs. "Transformer width $\times$1.5" (without domain labels) or "Tr. $\times$1.5 + domain tags" (with domain labels), both marked with $^*$; \underline{underline} denotes significant losses.}
\label{tab:ende_unseen_comet}
\end{table*}

% COMET ENFR SEEN
\begin{table*}[h!]
\normalsize
\centering
%\begin{tabular}{cp{1cm}p{1cm}p{1cm}p{1cm}p{1cm}p{1cm}}
\begin{tabular}{p{4.9cm}|ccccc|c}
\toprule
\textbf{Model} & \textbf{Law} & \textbf{Med} &   \textbf{Ted} & \textbf{Subt} & \textbf{Pat} &  \textbf{Avg} \\
\midrule
\multicolumn{7}{c}{Without access to seen domains labels} \\ \midrule
Transformer Base (TB)      &  89.6 &    86.7 &  81.5 &      76.9 &    77.7 &  82.5 \\
Transformer width x1.5     &  89.8 &    86.3 &  81.7 &      76.9 &    77.9 &  82.5 \\
Transformer width x5       &  90.1 &    86.6 &  82.1 &      77.3 &    78.0 &  82.8 \\
Mixture of Experts (SMoE)   &  90.0 &    86.4 &  81.8 &      77.1 &    77.8 &  82.6 \\
\midrule
\multicolumn{7}{c}{With access to seen domains labels} \\ \midrule
TB + domain tags           &  89.8 &    86.3 &  81.4 &      76.8 &    77.8 &  82.4 \\
TB + domain tags + DR      &  89.7 &    86.4 &  81.5 &      76.7 &    77.8 &  82.4 \\
Tr. x1.5 + domain tags     &  89.9 &    86.7 &  81.6 &      77.1 &    77.7 &  82.6 \\
Tr. 5 + domain tags + DR   &  90.0 &    86.6 &  82.1 &      77.3 &    78.1 &  82.8 \\
TB + domain adapters       &  89.9 &    86.2 &  79.9 &      76.7 &    77.7 &  82.1 \\
TB + domain adapters + DR  &  89.7 &    86.3 &  80.7 &      76.5 &    77.5 &  82.2 \\
MoE + domain tags          &  90.3 &    86.9 &  82.0 &      77.4 &    77.8 &  82.9 \\
MoE + domain tags + DR     &  90.3 &    86.7 &  81.9 &      77.4 &    78.0 &  82.8 \\
MoE + Dom. Aware Gate      &  90.1 &    86.7 &  81.4 &      77.1 &    77.8 &  82.6 \\
MoE + Dom. Aware Gate + DR &  90.1 &    86.5 &  81.9 &      77.2 &    78.0 &  82.7 \\
MoE + Dom. Spec. Gate      &  90.1 &    86.5 &  81.8 &      77.2 &    77.9 &  82.7 \\
MoE + Dom. Spec. Gate + DR &  90.2 &    86.4 &  81.9 &      77.5 &    77.9 &  82.8 \\
\bottomrule
\end{tabular}
\caption{COMET results for seen domains, English-French translation. DR: domain randomization.}
\label{tab:enfr_seen_comet}
\end{table*}

% COMET ENFR UNSEEN
\begin{table*}[ht!]
\normalsize
\centering
%\begin{tabular}{cp{1cm}p{1cm}p{1cm}p{1cm}p{1cm}p{1cm}}
\begin{tabular}{p{4.6cm}|cccp{1cm}p{1cm}p{1cm}c|c}
\toprule
\textbf{Model}   & \textbf{Flores} & \textbf{IT} & \textbf{Koran} & \textbf{Pat (med)} & \textbf{Subt (med)}& \textbf{World bank} & \textbf{ECB}  &  \textbf{Avg} \\
\midrule
\multicolumn{7}{c}{Without access to seen domains labels} \\ \midrule
Transformer Base (TB)      &   86.2 &  68.0 &  68.4 &      79.4 &       73.9 &      86.0 &  89.1 &  78.7 \\
Transformer width x1.5     &   86.2 &  68.0 &  69.2 &      79.4 &       74.2 &      86.1 &  89.3 &  78.9 \\
Transformer width x5       &   86.6 &  68.0 &  69.6 &      79.5 &       74.7 &      86.3 &  89.5 &  79.2 \\
Mixture of Experts (MoE)   &   86.3 &  68.0 &  69.5 &      79.5 &       74.2 &      86.2 &  89.2 &  79.0 \\
\midrule
\multicolumn{7}{c}{With access to seen domains labels} \\ \midrule
TB + domain tags           &   86.0 &  67.9 &  68.5 &      78.8 &       73.2 &      85.9 &  88.9 &  78.5 \\
TB + domain tags + DR      &   86.0 &  67.8 &  68.4 &      79.4 &       74.1 &      86.0 &  89.1 &  78.7 \\
Tr. x1.5 + domain tags     &   86.0 &  68.1 &  68.8 &      78.8 &       72.9 &      86.0 &  89.0 &  78.5 \\
Tr. 5 + domain tags + DR   &   86.6 &  68.3 &  70.0 &      79.6 &       74.5 &      86.2 &  89.2 &  79.2 \\
TB + domain adapters       &   86.0 &  68.1 &  69.0 &      76.8 &       71.8 &      85.8 &  88.7 &  78.0 \\
TB + domain adapters + DR  &   86.4 &  67.9 &  69.0 &      79.5 &       74.2 &      86.1 &  89.3 &  78.9 \\
MoE + domain tags          &   86.1 &  68.4 &  69.3 &      78.8 &       73.0 &      85.9 &  89.0 &  78.6 \\
MoE + domain tags + DR     &   86.5 &  68.1 &  69.5 &      79.5 &       74.4 &      86.2 &  89.4 &  79.1 \\
MoE + Dom. Aware Gate      &   86.2 &  68.2 &  69.1 &      78.0 &       72.7 &      86.0 &  89.0 &  78.5 \\
MoE + Dom. Aware Gate + DR &   86.4 &  68.2 &  69.1 &      79.5 &       74.3 &      86.0 &  89.4 &  79.0 \\
MoE + Dom. Spec. Gate      &   86.5 &  68.4 &  69.5 &      77.3 &       72.9 &      86.1 &  89.1 &  78.5 \\
MoE + Dom. Spec. Gate + DR &   86.6 &  68.1 &  69.2 &      79.5 &       74.5 &      86.3 &  89.3 &  79.1 \\
\bottomrule
\end{tabular}
\caption{COMET results for unseen domains, German-English translation. For models with access to seen domain labels (second group of rows), the ``generic'' label is used for unseen domains. DR: domain randomization.}
\label{tab:denfr_unseen_comet}
\end{table*}

\section{Analysis of gate statistics.}
\label{app:gate_stats}
To better assess difference between various gating mechanisms, we compare expert activations across different datasets of SMoE, SMoE+domain tags  and SMoE+domain aware gate, SMoE + Domain Spec gate with DR and without. 

\begin{figure*}
\includegraphics[width=\linewidth]{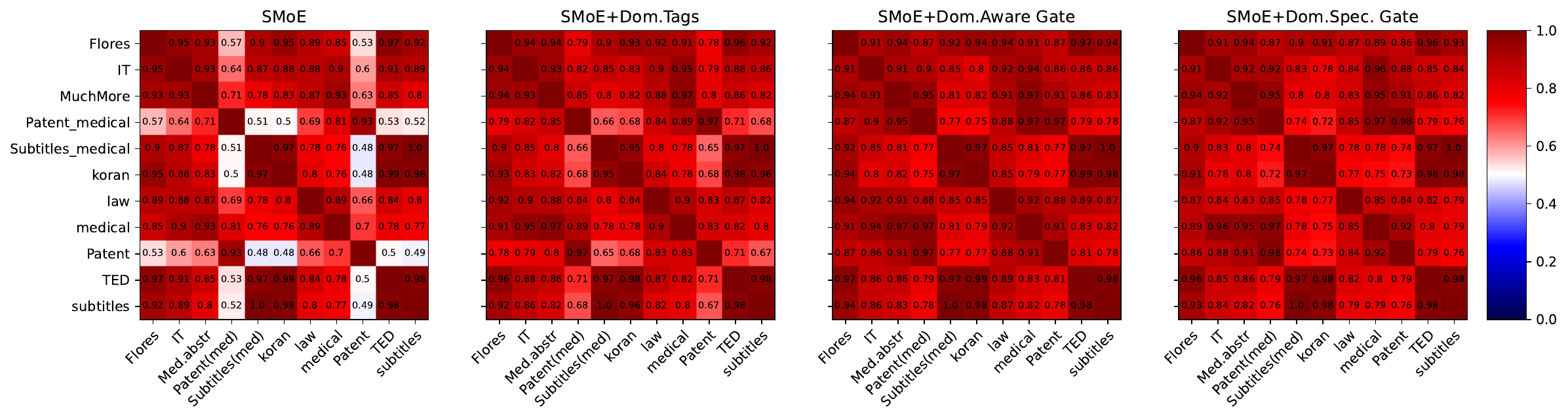}
\includegraphics[width=\linewidth]{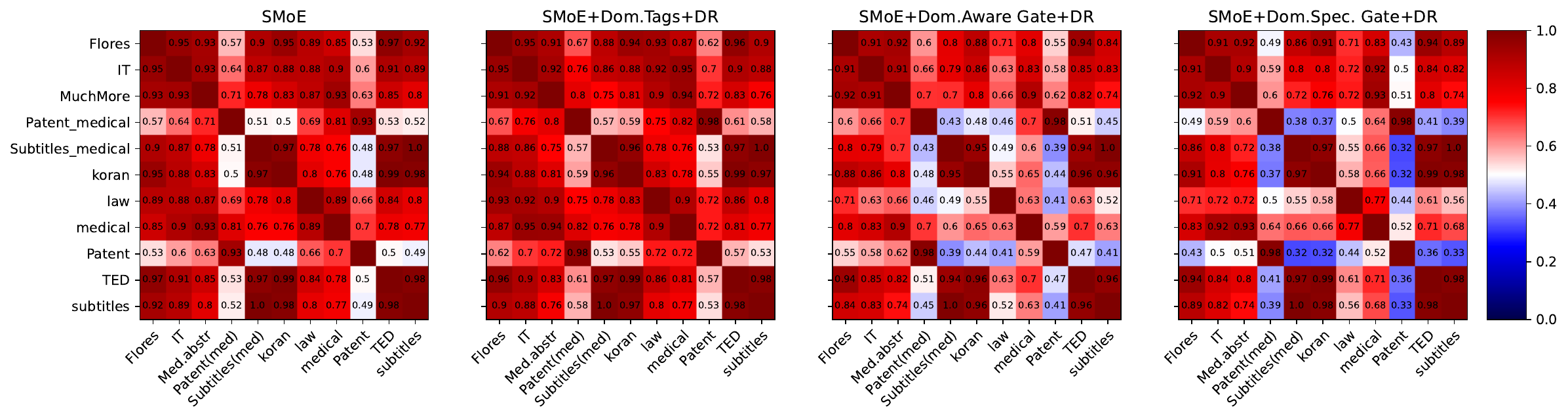}
    \caption{Pairwise similarity of experts activated across different datasets, De-En translation. All the datasets were decoded with the \textit{generic} domain embedding.}
    \label{fig:experts_spec_heatmap}
\end{figure*}

Figure \ref{fig:experts_spec_heatmap} plots the cosine similarity between the experts activation vectors for each pair of test sets. Following \cite{koishekenov2022memoryefficient} we compute the \textit{top-1 activity} metric for each expert as a fraction of tokens routed to this expert as the first choice. Each test set is then represented by a vector containing \textit{top-1 activities} of all the model's experts ($10$ experts $\times$ $6$ amount of layers with experts). The similarity between these vectors would indicate the similarities in experts' activation during decoding. 
 In this figure we use the \textit{generic} domain embedding for all the datasets during decoding. Therefore, it reflects how does an access to the domain knowledge \textit{during} training impacts models' capacity to specialize its experts based solely on token representation $x$. 

In the second analysis, we decode a single dataset (Subtitles)  with different domain labels. This allows to assess how the domain embedding impacts the choice of experts \textit{at inference}. Figure \ref{fig:experts_spec_by_dom_tag} plots similarities of the resulting experts' activation between different domain tags. 

We see that the models without DR rely much more on the domain tag (Figure \ref{fig:experts_spec_by_dom_tag}) compared to models with DR. Models with DR in its turn rely more on token representations to specialize experts compared to the model without DR (Figure \ref{fig:experts_spec_heatmap}). We note that models with Dom. Aware Gate and Dom. Specific Gate (right) lead to more dissimilar (and therefore more domain-specialized) experts overall compared to SMoE and SMoE+Domain.Tags models.  Model with no access to domain knowledge (SMoE) specializes its tokens more compared to models that have access to domain labels without DR (top part of the graph). 

%SMoE and SMoE+Dom.Tags + DR seem to behave similarly with slightly more specialization (lower similarity) for SMoE (which had no access to domain knowledge) for some domains (e.g. PatTR vs IT or TED). Based on these observations we can conclude that integration of domain knowledge via a prepended domain tag induces less expert specialization compared domain-aware gating or domain-specific gating. 

In addition, we note that all the models benefit from knowledge transfer by activating similar experts on similar domains: "Patents" and "Patents (medical)", "Subtitles" and "Subtitles (medical)", "Subtitles" and "Ted".

%We compute gate activation statistics for different models on each test dataset and compare them between the dataset. We rely on the experts \textit{top1 activity} metric, which compute the proportion of tokens for which each expert 

\end{document}